\documentclass[letterpaper, 10 pt, journal, twoside]{ieeetran}

\IEEEoverridecommandlockouts                              

\usepackage{graphicx} 
\usepackage{amsmath} 
\usepackage{amssymb}  

\usepackage{cite}

\usepackage[hidelinks]{hyperref}

\title{
A Hybrid Learning and Optimization Framework to Achieve Physically Interactive Tasks with Mobile Manipulators
}

\author{Jianzhuang Zhao$^{1,2,+}$, Alberto Giammarino$^{1,+}$, Edoardo Lamon$^{1}$, 
Juan M. Gandarias$^{1}$, \\
Elena De Momi$^{2}$, and Arash Ajoudani$^{1}$%
\thanks{Manuscript received: February, 24, 2022; Revised May, 15, 2022; Accepted June, 12, 2022.%Use only for final RAL version
This paper was recommended for publication by Editor Clement Gosselin upon evaluation of the Associate Editor and Reviewers' comments. 
This work was supported in part by the European Research Council's (ERC) starting grant Ergo-Lean (GA 850932), and in part by European Union’s Horizon 2020 research and innovation programme under Grant Agreement No. 871237 (SOPHIA). 
$^{+}$ Contributed equally to this work.}
\thanks{$^{1}$ Jianzhuang Zhao, Alberto Giammarino, Edoardo Lamon, Juan M. Gandarias, and Arash Ajoudani are with the Human-Robot Interfaces and physical Interaction lab, Istituto Italiano di Tecnologia, Genoa, Italy
        {\tt\footnotesize jianzhuang.zhao@iit.it; alberto.giammarino@iit.it; edoardo.lamon@iit.it; juan.gandarias@iit.it; arash.ajoudani@iit.it}.}%
\thanks{$^{2} $ Jianzhuang Zhao and Elena De Momi are with the Dept. of Electronics, Information and Bioengineering, Politecnico di Milano, Italy
        {\tt\footnotesize elena.demomi@polimi.it}.}
\thanks{Digital Object Identifier (DOI): 10.1109/LRA.2022.3187258.}
}

\begin{document}

\maketitle

\markboth{IEEE Robotics and Automation Letters. Preprint Version. Accepted June, 2022}
{Zhao, Giammarino \MakeLowercase{\textit{et al.}}: A Hybrid Learning and Optimization Framework} 

%%%%%%%%%%%%%%%%%%%%%%%%%%%%%%%%%%%%%%%%%%%%%%%%%%%%%%%%%%%%%%%%%%%%%%%%%%%%%%%%
\begin{abstract}

This paper proposes a hybrid learning and optimization framework for mobile manipulators for complex and physically interactive tasks.   
The framework exploits an admittance-type physical interface to obtain intuitive and simplified human demonstrations and Gaussian Mixture Model (GMM)/Gaussian Mixture Regression (GMR) to encode and generate the learned task requirements in terms of position, velocity, and force profiles. Next, using the desired trajectories and force profiles generated by GMM/GMR, the impedance parameters of a Cartesian impedance controller are optimized online through a Quadratic Program augmented with an energy tank to ensure the passivity of the controlled system. Two experiments are conducted to validate the framework, comparing our method with two approaches with constant stiffness (high and low). The results showed that the proposed method outperforms the other two cases in terms of trajectory tracking and generated interaction forces, even in the presence of disturbances such as unexpected end-effector collisions.
\end{abstract}

%%%%%%%%%%%%%%%%%%%%%%%%%%%%%%%%%%%%%%%%%%%%%%%%%%%%%%%%%%%%%%%%%%%%%%%%%%%%%%%%
\section{INTRODUCTION}
\IEEEPARstart{D}{ue} to the manufacturing shift from mass to customized production processes and the trend of an aging population in recent years, more flexibility and safety are required for robots in industrial and logistic scenarios. Collaborative Mobile Manipulators (CMM), which integrate a collaborative manipulator on a mobile platform, are designed with these objectives, combining the precision and manipulation capabilities of a robotic arm and the locomotion potential of the mobile platforms~\cite{roa2015mobile}. Because of these advantages, CMMs have demonstrated high suitability to different tasks, not only in industrial manufacturing and warehouse automation but also in service and domestic applications~\cite{roa2015mobile,bogh2014integration}.
Nevertheless, integrating different components makes the control and trajectory planning harder for CMM. Aiming at overcoming the synchronization issues of decoupled strategies, whole-body controllers have been presented. In particular, whole-body torque-based strategies could regulate not only the interaction forces with the environment  
but also the motion distribution at the joint level, generating different motion patterns
~\cite{lamon2020towards,wu2021unified}.

\begin{figure}[t]
    \centering
    \includegraphics[trim=0.0cm 0.0cm 0.0cm 0.0cm,clip,width=0.7\columnwidth]{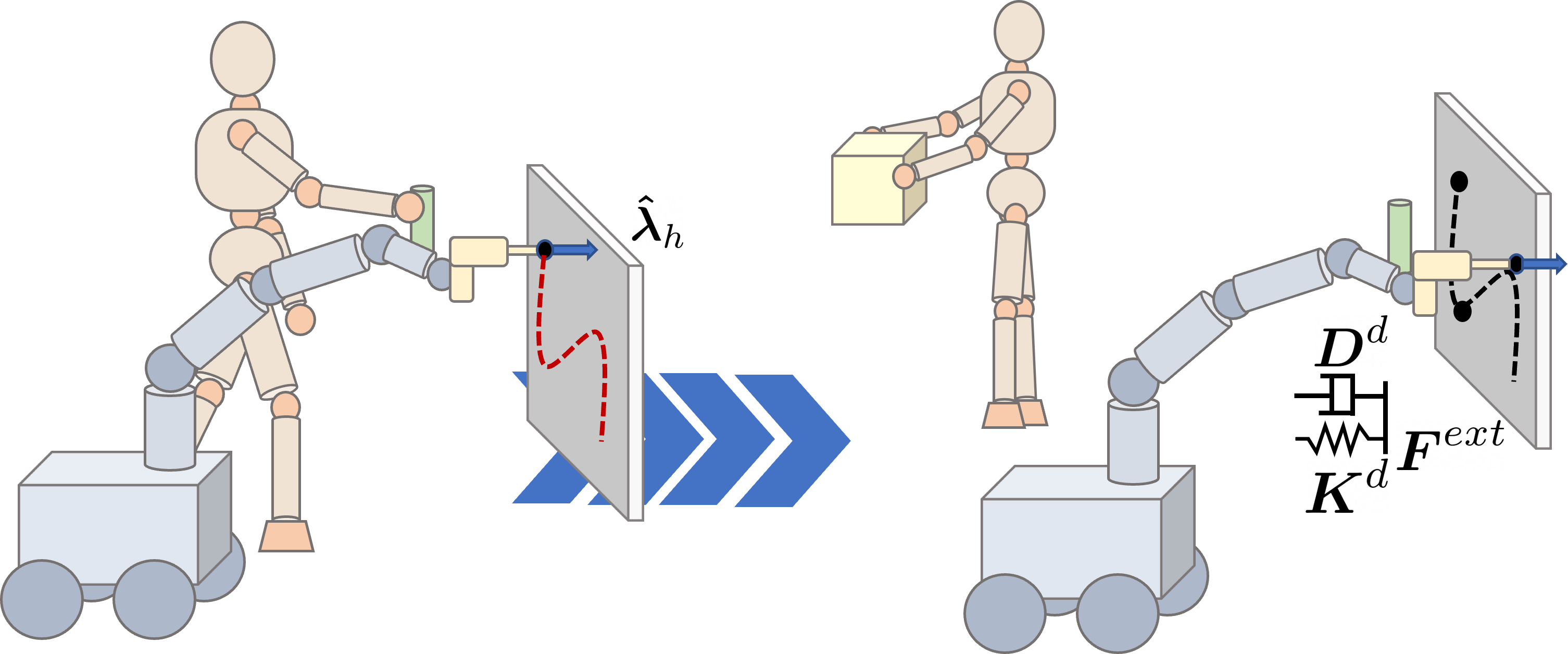} 
    \includegraphics[trim=0.0cm 0.0cm 0.0cm 0.0cm,clip,width=0.7\columnwidth]{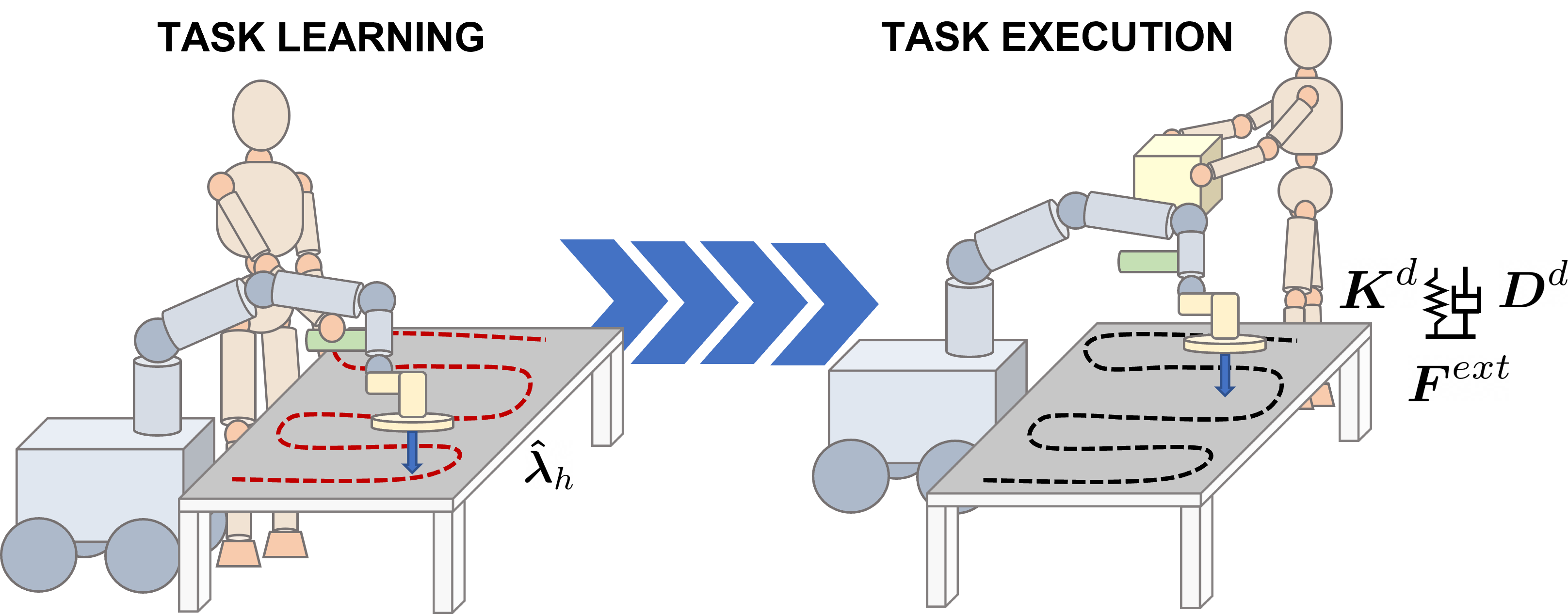}
    \caption{Left side: teaching a complex physically interactive task. Right side: Execution of the learned task adapting the Cartesian impedance.}
    \label{fig:intro}
    \vspace{-3mm}
\end{figure}

In addition, when facing unstructured environments, where interactions may change unexpectedly, collaborative robots (cobots) are expected to adapt their performance online to unforeseen situations.While dealing with physically interactive tasks (see Fig.~\ref{fig:intro}), hybrid position-force approaches are favorable w.r.t. pure force control for safety reasons. Nevertheless, such schemes presented an unstable behavior during the contact phase~\cite{averta2020enhancing}; hence it appears convenient to render virtual forces through an indirect method such as impedance control.

To model the system response to interactions with the environment, impedance controllers require the tuning of a high number of design parameters (generally stiffness, damping and inertia). Usually, high impedance is required in case of precise motions, whereas lower impedance is favorable in case of interactions with the environment.
However, how to optimally tune impedance parameters, namely variable impedance control (VIC), is still an open issue.
For instance, in \cite{duan2018adaptive} an impedance model based on a mass-damper system is used to express the desired dynamics of the error between the guessed position of the environment and the actual desired position, which is the one allowing force tracking. In particular, based on the force error the damping is tuned online and the desired position obtained from the resulting dynamic model is sent to a position-controlled robot. Although this method can achieve force tracking on different surfaces, the desired trajectory computed is altered based on the force error (e.g.,  when the surface contact is lost), which may result in an unsafe behavior of the robot.
Differently, authors in~\cite{averta2020enhancing, balatti2020method} propose different strategies to self-tune the impedance parameters according to the position error. While a high tracking error increases the stiffness, the interaction force is exploited to detect unplanned contacts and recover a compliant behavior.  In~\cite{averta2020enhancing}, moreover, the authors try to further minimize the interaction with the environment. Nevertheless, sometimes high interaction forces are a task requirement and hence becoming compliant is not a suitable strategy.  Additionally, human-in-the-loop approaches allow for online tuning of the robot behavior (varying damping and inertia) transferring the human skills for specific collaborative tasks as a trade-off between precision and speed of execution~\cite{ficucciello2015variable}. Other methods exploit EMG sensors to estimate joint stiffness and generate reference trajectories at the same time~\cite{peternel2014teaching,Wu2020a}.
Although these human impedance-transfer approaches for fixed base robotic arms can benefit from real-time adaptation to the changing environment, especially in teleoperation, EMG sensors and motion capture systems are required, making them unsuitable for industrial and domestic applications.

Furthermore, to generate autonomous tasks for CMM, optimal impedance profiles are not sufficient, also reference trajectories should be defined.
Imitation Learning (IL) paradigm has proven to be an efficient way to obtain desired motion given a predefined set of trajectories demonstrated by a teacher \cite{calinon2019learning}. The IL paradigm includes different algorithms to encode the taught demonstrations, such as Dynamical Movement Primitives (DMPs) \cite{ijspeert2013dynamical} and Gaussian Mixture Model (GMM) / Gaussian Mixture Regression (GMR) \cite{calinon2019learning}, among the most used. 
In the context of cobots, one IL methodology, i.e. kinesthetic teaching~\cite{calinon2019learning}, has gained increasing popularity since it does not require any retargeting from teacher to the robot. Its implementation usually requires gravity compensation or admittance control through force sensing (force-torque (F/T) sensors at the end-effector or joint torque sensors). 

To the best of the authors knowledge, currently there exist few intuitive interfaces that fully exploit the potential of CMMs in kinesthetic teaching. 
In the context of fixed-base manipulators, several works faced the problem of simultaneous learning of trajectories and force profiles through impedance models~\cite{abu2020variable} using GMM/GMR.
In \cite{calinon2010learning}, to generate safe robot behavior, the variable stiffness is tuned according the inverse of the position covariance of the GMM.
In \cite{abu2018forceras}, two different semi-positive definite (SPD) stiffness representations are investigated, while stiffness profiles are obtained with a least-squares estimator using a linear interaction model, based on position/force demonstrations. A similar approach based on regularized regression and force sensing was used by the authors in \cite{michel2021bilateral}, where the GMM maps the relationship between external force (input) and stiffness (output). Then, the stiffness is generated online by GMR based on current external force during teleoperation, where high external force will result in high stiffness. Nonetheless, the high stiffness of robots may hurt humans if the large force is generated by human disturbances. A different methodology, based on reinforcement learning (RL), models the robot policy by DMP that includes trajectory and impedance parameters. The policy parameters are then offline optimized using a variant of policy improvement with path integrals~\cite{buchli2011learning}.
However, this RL method and the iteration optimization method in \cite{averta2020enhancing} requires several simulated executions of the desired task and they cannot deal with unforeseen disturbances online.
In addition, most of the methods in the literature do not directly ensure the stability of the VIC, whose property is fundamental for a safe Human-Robot Collaboration (HRC)~\cite{abu2020variable}.

\begin{figure*}[t]
	\centering
	\includegraphics[trim=0cm 0cm 0cm 0cm,clip,width=0.9\linewidth]{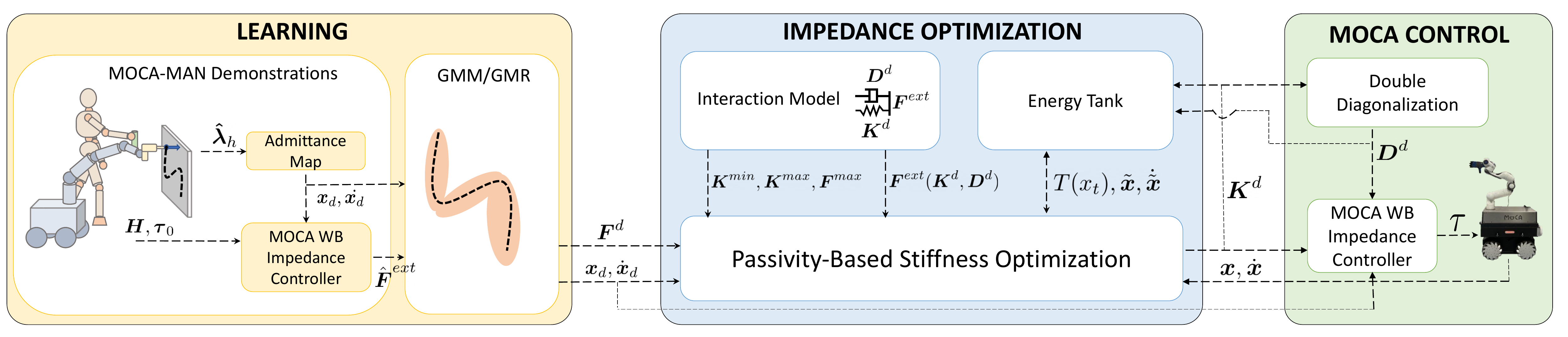}
	\caption{The proposed framework scheme. Humans can teach MOCA the interactive tasks directly through an admittance-type physical interface, and the desired trajectories and interaction force are replicated with GMM/GMR. Then, the desired force is sent to a QP-based algorithm to optimize online stiffness. A tank energy constraint ensures the passivity of the system. Finally, the desired trajectory and stiffness are sent to MOCA's whole-body impedance controller.}
	\label{fig:framework}
\end{figure*}

 To address these issues, we propose a method to learn loco-manipulation tasks exploiting the paradigm of VIC %variable impedance control 
(see Fig.~\ref{fig:intro}). Our goal is to teach our MObile Collaborative robotic Assistant (MOCA) from human physical demonstrations through IL and generate robust and safe autonomous behaviors by regulating the interaction forces between the robot and the unstructured environment in quasi-static conditions.
Thanks to an admittance-type interface~\cite{gandarias2022enhancing}, complex interactive tasks are intuitively demonstrated and then encoded by a GMM that learns the task trajectory and interaction force maps between the robot and the environment simultaneously. The desired trajectory and force profile are generated using GMR. Based on the learned force, the desired impedance parameters are computed online as the result of a quadratic program (QP), which ensures the lowest stiffness required to achieve the task, subject to force and stiffness limits imposed by safety requirements.
Moreover, an energy tank-based passivity constraint is added to ensure the controller's stability. The desired trajectory and generated stiffness are finally sent to the MOCA weighted whole-body impedance controller (see Fig.~\ref{fig:framework}). 

The method is evaluated with a table cleaning task, where two different experiments are conducted. First of all, the task is taught in nominal conditions. Then, MOCA replicates the learned task in the same conditions and in the presence of external disturbances coming from the environment and from unexpected physical interactions with the human. The results show that the impedance parameters' tuning yields similar (good) performances in all conditions, allowing at the same time trajectory and force tracking and robust and compliant interactions compared to constant stiffness solutions.

\section{PRELIMINARIES}

\subsection{MOCA Platform and Whole-Body Impedance Control}
\label{subsec:MOCA_wb_controller}
MOCA is a research robotic platform designed for HRC, with loco-manipulation capabilities that make it potentially suitable for logistics~\cite{lamon2020towards}  
and flexible manufacturing~\cite{fusaro2020human}. 
It is composed of the lightweight torque-controlled 7-DoFs Franka Emika Panda robotic arm, mounted on top of the velocity-controlled 3-DoFs Robotnik SUMMIT-XL STEEL mobile platform.

The whole-body dynamic model~\cite{dietrich2016whole, wu2021unified} can be formulated as:
\begin{equation}
\begin{aligned}
  & 
  \begin{pmatrix} 
  \boldsymbol{M}_{v} & \boldsymbol{0} \\ \boldsymbol{0} & \boldsymbol{M}_{a} (\boldsymbol{q}_a) \end{pmatrix} 
  \begin{pmatrix}
  \ddot{\boldsymbol{q}}_m \\ \ddot{\boldsymbol{q}}_a
  \end{pmatrix} + 
  \begin{pmatrix} 
  \boldsymbol{D}_{v} & \boldsymbol{0} \\ \boldsymbol{0} & \boldsymbol{C}_{a} (\boldsymbol{q}_a, \dot{\boldsymbol{q}_a}) \end{pmatrix}  
  \begin{pmatrix}
  \dot{\boldsymbol{q}}_m \\ \dot{\boldsymbol{q}}_a
  \end{pmatrix} \\
  &+
  \begin{pmatrix}
  \boldsymbol{0} \\ \boldsymbol{g}_{a} (\boldsymbol{q}_a)
  \end{pmatrix}
   =
  \begin{pmatrix}
  \boldsymbol{\tau}_m^{vir} \\ \boldsymbol{\tau}_{a}
  \end{pmatrix} +
  \begin{pmatrix}
  \boldsymbol{\tau}_m^{ext} \\ \boldsymbol{\tau}_{a}^{ext}
  \end{pmatrix},
  \label{eq:whole_body_dynamics_extended}
\end{aligned}
\end{equation}

\noindent where $\boldsymbol{M}_{v} \in\mathbb{R}^{n_b \times n_b}$ and $\boldsymbol{D}_{v} \in\mathbb{R}^{n_b \times n_b}$ are the virtual inertia and virtual damping of the mobile platform, $\boldsymbol{\dot{q}}_{m} \in\mathbb{R}^{n_b}$ is its input velocity, $\boldsymbol{\tau}_{m}^{ext} \in\mathbb{R}^{n_b}$ and $\boldsymbol{\tau}_{m}^{vir} \in\mathbb{R}^{n_b}$ are the related external and the virtual torque. With respect to the arm, $\boldsymbol{q}_{a}, \boldsymbol{\dot{q}}_{a},$ and $\boldsymbol{\ddot{q}}_{a} \in\mathbb{R}^{n_a}$ are the joint angles, velocities and accelerations vectors, $\boldsymbol{M}_{a} \in\mathbb{R}^{n_a \times n_a}$ is the symmetric and positive definite inertia matrix of the arm, $\boldsymbol{C}_{a} \in\mathbb{R}^{n_a}$ is the Coriolis and centrifugal force vector, $\boldsymbol{g}_{a} \in \mathbb{R}^{n_a}$ is the gravity, $\boldsymbol{\tau}_{a} \in\mathbb{R}^{n_a}$, and $\boldsymbol{\tau}_{a}^{ext} \in\mathbb{R}^{n_a}$ are the commanded torque vector and external torque vector, respectively.
This model can be summarized by 
\begin{equation} \label{eq:whole_body_dynamics}
    \boldsymbol{M}(\boldsymbol{q}) \ddot{\boldsymbol{q}} + \boldsymbol{C}(\boldsymbol{q},\dot{\boldsymbol{q}})\dot{\boldsymbol{q}} + \boldsymbol{g}(\boldsymbol{q}) = \boldsymbol{\tau} + \boldsymbol{\tau}^{ext}.
\end{equation}
where $\boldsymbol{M}(\boldsymbol{q}) \in\mathbb{R}^{n \times n}$ ($n=n_a+n_b$) is the symmetric positive definite joint-space inertia matrix, $\boldsymbol{C}(\boldsymbol{q},\dot{\boldsymbol{q}}) \in\mathbb{R}^{n \times n}$ is the joint-space Coriolis/centrifugal matrix, and $\boldsymbol{g}(\boldsymbol{q})\in\mathbb{R}^n$ the joint-space gravity. Finally, $\boldsymbol{\tau}\in\mathbb{R}^{n}$ and $\boldsymbol{\tau}^{ext}\in\mathbb{R}^{n}$ represent joint-space input and external torque.

The MOCA Cartesian impedance controller is formulated as a prioritized weighted inverse dynamics algorithm and can be obtained by solving the problem of finding the torque vector $\boldsymbol{\tau}$ closest to some desired $\boldsymbol{\tau_0}$ that realizes the operational forces $\boldsymbol{F}\in\mathbb{R}^m$ ($m \le 6$), according to the norm induced by the positive definite weighting matrix $\boldsymbol{W} \in\mathbb{R}^{n \times n}$, 
\begin{equation}\label{eq:control_opt_problem}
    \underset{\boldsymbol{\tau \in\mathbb{R}^{n}}}{\min} \: \frac{1}{2} \left\| \boldsymbol{\tau} - \boldsymbol{\tau_0} \right\|_{\boldsymbol{W}}^2 
    \quad  \text{s.t.   } \boldsymbol{F} = \bar{\boldsymbol{J}}^{T}\boldsymbol{\tau}
\end{equation}
where 
% \begin{equation}\label{eq:dynamic_jacobian} 
    $\bar{\boldsymbol{J}} = \boldsymbol{M}^{-1} \boldsymbol{J}^{T} \boldsymbol{\Lambda}$
% \end{equation}
is the dynamically consistent pseudo-inverse of the Jacobian matrix $\boldsymbol{J}(\boldsymbol{q})$, and the constraint
% \begin{equation} \label{eq:force_torque_rel}
$
    \boldsymbol{F} = \bar{\boldsymbol{J}}^{T} \boldsymbol{\tau} = \boldsymbol{\Lambda}\boldsymbol{J}\boldsymbol{M}^{-1} \boldsymbol{\tau}
$
% \end{equation}
is the general relationship between the generalized joint torques and the operational forces, and $\boldsymbol{\Lambda} = { ( \boldsymbol{J}\boldsymbol{M}^{-1} \boldsymbol{J}^{T})}^{-1} \in\mathbb{R}^{m \times m}$ is the Cartesian inertia. 
The closed-form solution results in:   
\begin{equation}
\begin{aligned} \label{eq:opt_solution}
    \boldsymbol{\tau} = & \boldsymbol{W}^{-1} \boldsymbol{M}^{-1} \boldsymbol{J}^{T} \boldsymbol{\Lambda}_{W} \boldsymbol{\Lambda}^{-1} \boldsymbol{F} +\\
    &+ (\boldsymbol{I}-\boldsymbol{W}^{-1} \boldsymbol{M}^{-1} \boldsymbol{J}^{T} \boldsymbol{\Lambda}_{W} \boldsymbol{J} \boldsymbol{M}^{-1})\boldsymbol{\tau}_0 ,
\end{aligned}
\end{equation}
where $\boldsymbol{\Lambda_{W}} = ( \boldsymbol{J} \boldsymbol{M}^{-1} \boldsymbol{W}^{-1} \boldsymbol{M}^{-1} \boldsymbol{J}^{T} )^{-1} $ is the weighted Cartesian inertia, analogous to the Cartesian inertia $\boldsymbol{\Lambda}$.

The weighting matrix $\boldsymbol{W}$ is generally defined as $\boldsymbol{W}(\boldsymbol{q})=\boldsymbol{H}^T\boldsymbol{M}^{-1}(\boldsymbol{q})\boldsymbol{H}$, where $\boldsymbol{H}\in\mathbb{R}^{n \times n}$ is the tunable positive definite weight matrix of the controller, that is used to generate different motion modes~\cite{lamon2020towards}.

Finally, $\boldsymbol{F}$ is computed to generate the desired closed-loop behaviour, according to the Cartesian impedance law:
\begin{equation}
\boldsymbol{F} =  - \boldsymbol{D}^d\dot{\boldsymbol{x}} - \boldsymbol{K}^d\Tilde{\boldsymbol{x}},
    \label{eq:cartesian_impedance}
\end{equation}
where $\Tilde{\boldsymbol{x}} = \boldsymbol{x}_d - \boldsymbol{x} \in\mathbb{R}^{6}$ is the Cartesian pose error computed with respect to the desired Cartesian equilibrium pose $\boldsymbol{x}_d$, and $\boldsymbol{D}^d$, $\boldsymbol{K}^d \in\mathbb{R}^{m \times m}$ are the desired Cartesian damping and stiffness matrices, respectively.
Moreover, $\boldsymbol{\tau_0}$ could contain different contributions, such as joint impedance, collision and self-collision avoidance, joint limits avoidance, etc.

\subsection{GMM \& GMR}
The desired trajectories and force profiles are generated using GMM and GMR. This choice is due to the low number of hyperparameters that need to be tuned (i.e. only $K$, the number of Gaussians).
The parameters of GMM can be estimated by Expectation-Maximization (EM) algorithm~\cite{bishop2006pattern} with an offline training process that makes use of the demonstrations. 
To make the representation easier to understand, we give the following definition:  $\boldsymbol{\eta}^{\mathcal{I}}$ and $\boldsymbol{\eta}^{\mathcal{O}}$ respectively denote the input and output variables on which the training is carried out, where the superscripts $\mathcal{I}$ and $\mathcal{O}$ stand for their dimensions, respectively. 
Given a input variable $\boldsymbol{\eta}^{\mathcal{I}}$, the best estimation of output $\boldsymbol{\hat{\eta}}^{\mathcal{O}}$ is the mean $\boldsymbol{\hat{\mu}}$ of the conditional
probability distribution $\boldsymbol{\hat{\eta}}^\mathcal{O}|\boldsymbol{\eta}^\mathcal{I} \sim \mathcal{N}(\boldsymbol{\hat{\mu}}, \boldsymbol{\hat{\Sigma}})$, which is computed by GMR \cite{calinon2019learning}.

\section{METHODOLOGY}
\subsection{Desired Trajectory and Interaction Force Hybrid Learning Exploiting an admittance-type physical interface.}
\label{subsec:kinesthetic_teaching}

Aiming at demonstrating desired end-effector trajectories and interaction wrenches to MOCA, we integrated kinesthetic teaching with an admittance-type physical interface. 
The interface is depicted in Fig.~\ref{fig:experiment_setup}. It consists of i) an Arduino Nano microcontroller connected to a 4-buttons panel that allows the user to configure different functionalities and communicate
with the robot through the Robot Operating System (ROS) middleware suite, ii) a F/T sensor to measure the user interaction wrenches with a physical part that the human can easily grasp and iii) an end-effector tool. A detailed explanation of the hardware and design of an admittance-type physical interface can be found in~\cite{gandarias2022enhancing}. 

The measured human wrenches $\boldsymbol{\hat{\lambda}}_h$ $\in \mathbb{R}^m$, are used to change the input Cartesian equilibrium pose in \eqref{eq:cartesian_impedance} as input of an admittance controller which implements the following desired dynamics:

\begin{equation}
\label{eq:standard_adm_controller}
    \boldsymbol{M}^{adm}\ddot{\boldsymbol{x}}_d + \boldsymbol{D}^{adm}\dot{\boldsymbol{x}}_d = \boldsymbol{\hat{\lambda}}_h,
\end{equation}
where $\boldsymbol{M}^{adm},\boldsymbol{D}^{adm}\in \mathbb{R}^{m\times m}$ are respectively the admittance mass and damping. 
This way, MOCA can move in space in the same direction of the applied wrench.  
During the demonstration, the aforementioned buttons enable the human teacher to control some functionalities of the admittance mapping in \eqref{eq:standard_adm_controller} and of the whole-body controller presented in section \ref{subsec:MOCA_wb_controller}. In particular, the human has control over the loco-manipulation behavior of the robot (through selection of $\boldsymbol{\tau_0}$ and $\boldsymbol{H}$) and over the desired admittance ($\boldsymbol{M}^{adm}$ and $\boldsymbol{D}^{adm}$). In addition, the human can also enable and disable rotations and translations of the end-effector. 

The additional features introduced by an admittance-type physical interface allow to exploit the full potential of a mobile base robot and ease the demonstration process, considering the human preference in different parts of the demonstrated behavior. Thanks to selecting the loco-manipulation mode, the human teacher can move the robot in a theoretically infinite workspace during the demonstration and keep the mobile base still when an interaction of the end-effector with the environment is planned. Indeed, since the mobile base movements, controlled at lower frequency w.r.t. the arm, perturb the end-effector motion, demonstrating a desired interaction with the environment can result challenging.  
Then, the human teacher can tune the system's responsiveness to the interaction by selecting the admittance according to his/her experience, preference, and specific requirements of the task: high admittance for long unconstrained motion and low admittance for short and precise interactions with the environment. Finally, selecting only a subset of the task space axes could help execute the task and speed up the learning procedure.

Desired Cartesian equilibrium pose $\boldsymbol{x}_d$ and twist $\boldsymbol{\dot{x}}_d$ are obtained as the solution of \eqref{eq:standard_adm_controller}. Since the robotic arm used is torque controlled, the interaction wrenches of the robot end-effector with the environment $\hat{\boldsymbol{F}}^{ext}$ can be estimated through the joint torque measurements of the arm. 
Hence, during the demonstration, the human teacher can guide the robot through the admittance-type physical interface while the robot arm estimates its interaction with the environment. Then, a GMM is trained from the data obtained from the demonstrations, which input variable is time  $\boldsymbol{\eta}^{\mathcal{I}} = t$ and output variables are $\boldsymbol{\eta}^{\mathcal{O}} = \begin{bmatrix} \boldsymbol{x}_d^T & \boldsymbol{\dot{x}}_d^T & {\hat{\boldsymbol{F}}^{ext^T}} \end{bmatrix}^T$.  
Finally, the desired trajectory and force can be generated by GMR.

\subsection{QP-Based Stiffness Online Optimization}
\label{subsec:QP_formulation}
The outputs of the GMR are fed to a QP that allows to online modulate the stiffness of the Cartesian whole-body impedance controller presented in section \ref{subsec:MOCA_wb_controller}. The QP is formulated as follows:
\begin{align}
    \min_{\substack{\boldsymbol{K}_{i}^d \in \mathbb{R}^{m \times m} \\ i \in \{1,\dots,N\}}} \: & \frac{1}{2} \sum_{i=1}^{N} \left( \| \boldsymbol{F}_{i}^{ext} - \boldsymbol{F}_{i}^{d} \|_{\boldsymbol{Q}}^2 + \| \boldsymbol{K}^{d}_i - \boldsymbol{K}^{min} \|_{\boldsymbol{R}}^2 \right) \nonumber \\ 
    \quad  \text{s.t.   } \qquad &  \boldsymbol{K}^{min} \le \boldsymbol{K}_{i}^d \le \boldsymbol{K}^{max} \hspace{0.5cm} i \in \{1,\dots,N\} \label{eq:general_QP_formulation} \\
    -&\boldsymbol{F}^{max} \le \boldsymbol{F}_{i}^{ext} \le \boldsymbol{F}^{max} \hspace{0.4cm} i \in \{1,\dots,N\} \nonumber 
\end{align}
where $i$ is the time step, $N$ is the length of the time window, $\boldsymbol{Q}$ and $\boldsymbol{R}$ $\in \mathbb{R}^{m\times m}$ are diagonal positive definite weighting matrices, $\boldsymbol{K}^d_i \in \mathbb{R}^{m\times m}$ is the desired stiffness of the Cartesian whole body impedance controller at time step $i$, $\boldsymbol{K}^{min}$ and $\boldsymbol{K}^{max}$ $\in \mathbb{R}^{m\times m}$ are respectively minimum and maximum allowed stiffness, $\boldsymbol{F}^{ext}_i \in \mathbb{R}^{m}$ is the wrench of the impedance interaction model at time step $i$, that can be modeled with impedance-like laws, $\boldsymbol{F}^d_i \in \mathbb{R}^{m}$ is the learned desired interaction wrench at time step $i$ (i.e., output of the GMR) and $\boldsymbol{F}^{max} \in \mathbb{R}^{m}$ is the maximum wrench that the robot can exert. The constraint inequality between vectors is element-wise. The optimization problem formulated above trades off the tracking of the desired wrench 
with the requirement of keeping a small stiffness.

The desired impedance interaction model can be expressed in different possible ways. To render a desired mass-spring-damper system, $\boldsymbol{F}^{ext}$ is expressed as:
\begin{equation}
\boldsymbol{F}^{ext} = \boldsymbol{\Lambda}^d\ddot{\Tilde{\boldsymbol{x}}} + \boldsymbol{D}^d\dot{\Tilde{\boldsymbol{x}}} + \boldsymbol{K}^d\Tilde{\boldsymbol{x}},
    \label{eq:cartesian_impedance_inertia_shaping}
\end{equation}
where $\boldsymbol{\Lambda}^d \in \mathbb{R}^{m \times m}$ is the desired inertia matrix. However, the interaction wrench must be measured precisely to render the desired inertia matrix, which is referred to as inertia shaping (e.g., using an F/T sensor). Unfortunately, a precise measure is often not available, and the following interaction model is rendered \cite{ott2008cartesian}:
\begin{equation}
     \boldsymbol{F}^{ext} =
     \boldsymbol{\Lambda}(\boldsymbol{x}) \ddot{\Tilde{\boldsymbol{x}}} + \big( \boldsymbol{\mu}(\boldsymbol{x},\dot{\boldsymbol{x}}) + \boldsymbol{D}^d \big) \dot{\Tilde{\boldsymbol{x}}} + \boldsymbol{K}^d\Tilde{\boldsymbol{x}},
    \label{eq:cartesian_impedance_robot_inertia}
\end{equation}
where no inertia shaping is performed, since the actual Cartesian inertia of the manipulator $\boldsymbol{\Lambda}(\boldsymbol{x})$ is used. 
Note that, in order to keep the physical coherence of the interaction model, the Coriolis and centrifugal terms $\boldsymbol{\mu}(\boldsymbol{x},\dot{\boldsymbol{x}})$ must be added to the desired damping since those arise from a configuration dependent inertia~\cite{ott2008cartesian, ficucciello2015variable}.

In practice, since also the acceleration signal is often noisy, the following simplified model is used:
\begin{equation}
\boldsymbol{F}^{ext} = \big( \boldsymbol{\mu}(\boldsymbol{x},\dot{\boldsymbol{x}}) + \boldsymbol{D}^d \big)\dot{\Tilde{\boldsymbol{x}}} + \boldsymbol{K}^d\Tilde{\boldsymbol{x}}.
    \label{eq:cartesian_impedance_no_inertia}
\end{equation}

\subsection{Tank Energy Based Passivity Constraint}
The QP formulated in section \ref{subsec:QP_formulation} computes a Cartesian stiffness at each time step that is sent to the Cartesian impedance whole-body controller presented in section \ref{subsec:MOCA_wb_controller}. It is well known that VIC might violate the passivity of the system~\cite{ferraguti2015energy}; thus, the stability of the controlled system is not guaranteed. In order to ensure system stability, the QP formulation previously presented can be augmented through a passivity constraint for the power port $\dot{\boldsymbol{x}}\boldsymbol{F}^{ext}$. 
The interaction model of the variable Cartesian impedance can be written as a port-Hamiltonian system and augmented through an energy tank \cite{ferraguti2015energy}. The scalar differential equation that describes the tank dynamics is: 
\begin{equation}
    \dot{\textrm{x}}_t = \frac{\sigma}{\textrm{x}_t}\dot{\Tilde{\boldsymbol{x}}}^T\boldsymbol{D}^d\dot{\Tilde{\boldsymbol{x}}} - \frac{\boldsymbol{w}^T}{\textrm{x}_t}\dot{\Tilde{\boldsymbol{x}}},
\end{equation}
where $\textrm{x}_t \in \mathbb{R}$ is the state of the tank that stores energy $T(\textrm{x}_t)=\frac{1}{2}\textrm{x}_t^2$,  $\sigma \in \{0,1\}$ is used to enable and disable the dissipated energy storage in case a maximum limit is reached, and $\boldsymbol{w}$ is the extra input of the port-Hamiltonian dynamics which can be written as:
\begin{equation}
    \boldsymbol{w}(t) = \begin{cases} -\boldsymbol{K}^v(t)\Tilde{\boldsymbol{x}} & \mbox{if $T(\textrm{x}_t)>\varepsilon$}  \\ 0 & \mbox{otherwise,} \end{cases}
\end{equation}
where $\boldsymbol{K}^v(t)$ is the variable part of the stiffness so that $\boldsymbol{K}^d(t) = \boldsymbol{K}^{min} + \boldsymbol{K}^v(t)$ and $\varepsilon>0$ is the minimum energy that the tank is allowed to store. The tank energy is initialized so that $T(\textrm{x}_t(0))>\varepsilon$.
The constraint used to augment the QP is derived by integrating the energy tank over time and enforcing it to be higher than its minimum $\varepsilon$:
\begin{equation}
   T(\textrm{x}_t(t)) = T(\textrm{x}_t(t-1)) + \dot{T}(\textrm{x}_t(t))\Delta t > \varepsilon,
\end{equation}
where $\Delta t$ is the time step and
\begin{align}
    \dot{T}(\textrm{x}_t) &= \textrm{x}_t\dot{\textrm{x}}_t = \sigma\dot{\Tilde{\boldsymbol{x}}}^T\boldsymbol{D}^d\dot{\Tilde{\boldsymbol{x}}} - \boldsymbol{w}^T\dot{\Tilde{\boldsymbol{x}}} \\
    &= \begin{cases} \sigma\dot{\Tilde{\boldsymbol{x}}}^T\boldsymbol{D}^d\dot{\Tilde{\boldsymbol{x}}} + \Tilde{\boldsymbol{x}}^T\boldsymbol{K}^v\dot{\Tilde{\boldsymbol{x}}} & \mbox{if $T(\textrm{x}_t)>\varepsilon$} \\
    \sigma\dot{\Tilde{\boldsymbol{x}}}^T\boldsymbol{D}^d\dot{\Tilde{\boldsymbol{x}}} & \mbox{otherwise.}\end{cases}
    \label{eq:Tank_derivative}
\end{align}
Note from \eqref{eq:Tank_derivative} that the constraint can be expressed as a linear function of the QP optimization variable only when $T>\varepsilon$. Instead, when $T \le \varepsilon$ the stiffness is constrained to take on its minimum value ($\boldsymbol{K}^d = \boldsymbol{K}^{min}$). 

\subsection{Technical Details}
While in the previous subsection, the method has been explained with generality, some simplifying hypotheses have been assumed in the implemented method. First of all, in the QP, the interaction model in \eqref{eq:cartesian_impedance_no_inertia} is used, where $\boldsymbol{K}^{d}, \boldsymbol{D}^{d}$ are assumed diagonal and $\dot{\boldsymbol{x}}_d$ is assumed equal to zero to comply with the whole-body controller formulation. Moreover, only the translational part of the interaction model is considered in the optimization, while the rotational part is kept constant ($m = 3$). The value of $\boldsymbol{D}^{d}$ in the optimization at the next control loop is computed through double diagonalization at the previous optimization step $\boldsymbol{K}^{d}$, with critical damping factor~\cite{ott2008cartesian}, where the diagonal elements are:
\begin{equation}
    \boldsymbol{d}^d(t) = 2 \cdot 0.707 \cdot \sqrt{\boldsymbol{k}^d(t-1)},
    \label{eq:double_diag}
\end{equation}
where $\boldsymbol{d}^d(t) \in \mathbb{R}^3$ is the vector of the diagonal components of the desired Cartesian damping matrix at time $t$ and $\boldsymbol{k}^d(t-1) \in \mathbb{R}^3$ is the vector of the diagonal components of the desired Cartesian stiffness matrix at time $t-1$. The damping is computed using the desired stiffness at the previous time step to preserve the cost function's quadratic nature in the QP formulation. Finally, only a one-time step is considered in each optimization ($N=1$).

The problem in \eqref{eq:general_QP_formulation} can be rewritten as:
\begin{align}
    \min_{\substack{\boldsymbol{k}^d \in \mathbb{R}^3}} \: & \frac{1}{2} \left( \| \boldsymbol{F}^{ext} - \boldsymbol{F}^{d} \|_{\boldsymbol{Q}}^2 + \| diag\{\boldsymbol{k}^{d}\} - \boldsymbol{K}^{min} \|_{\boldsymbol{R}}^2 \right) \nonumber \\
    \quad  \text{s.t.   } \qquad &  \boldsymbol{k}^{min} \le \boldsymbol{k}^d \le \boldsymbol{k}^{max} \label{eq:specific_QP_formulation} \\
    -&\boldsymbol{F}^{max} \le \boldsymbol{F}^{ext} \le \boldsymbol{F}^{max} \nonumber \\ 
    & T(\textrm{x}_t) \ge \varepsilon \nonumber
\end{align}
where $diag\{\cdot\}$ is the diagonal operator and the vector inequalities are elementwise. Note that both $\boldsymbol{F}^{ext}$ and $T(\textrm{x}_t)$ are linear functions of the optimization variable $\boldsymbol{k}^d$, thus the last two inequality constraints can be easily expressed in the generic form $\boldsymbol{C}\boldsymbol{k}^d \le \boldsymbol{d}$. Moreover, if $T(\textrm{x}_t) < \varepsilon$, then $\boldsymbol{k}^d = \boldsymbol{k}^{min}$. Note also that, since only the translational part of the impedance model is considered, the GMM is trained to encode only the linear part of velocities and forces.

\section{EXPERIMENTS AND RESULTS}
\subsection{Experimental Setup}

\begin{figure}[t]
	\centering
    \includegraphics[trim=0cm 0cm 0cm 0cm,clip,width=0.75\linewidth]{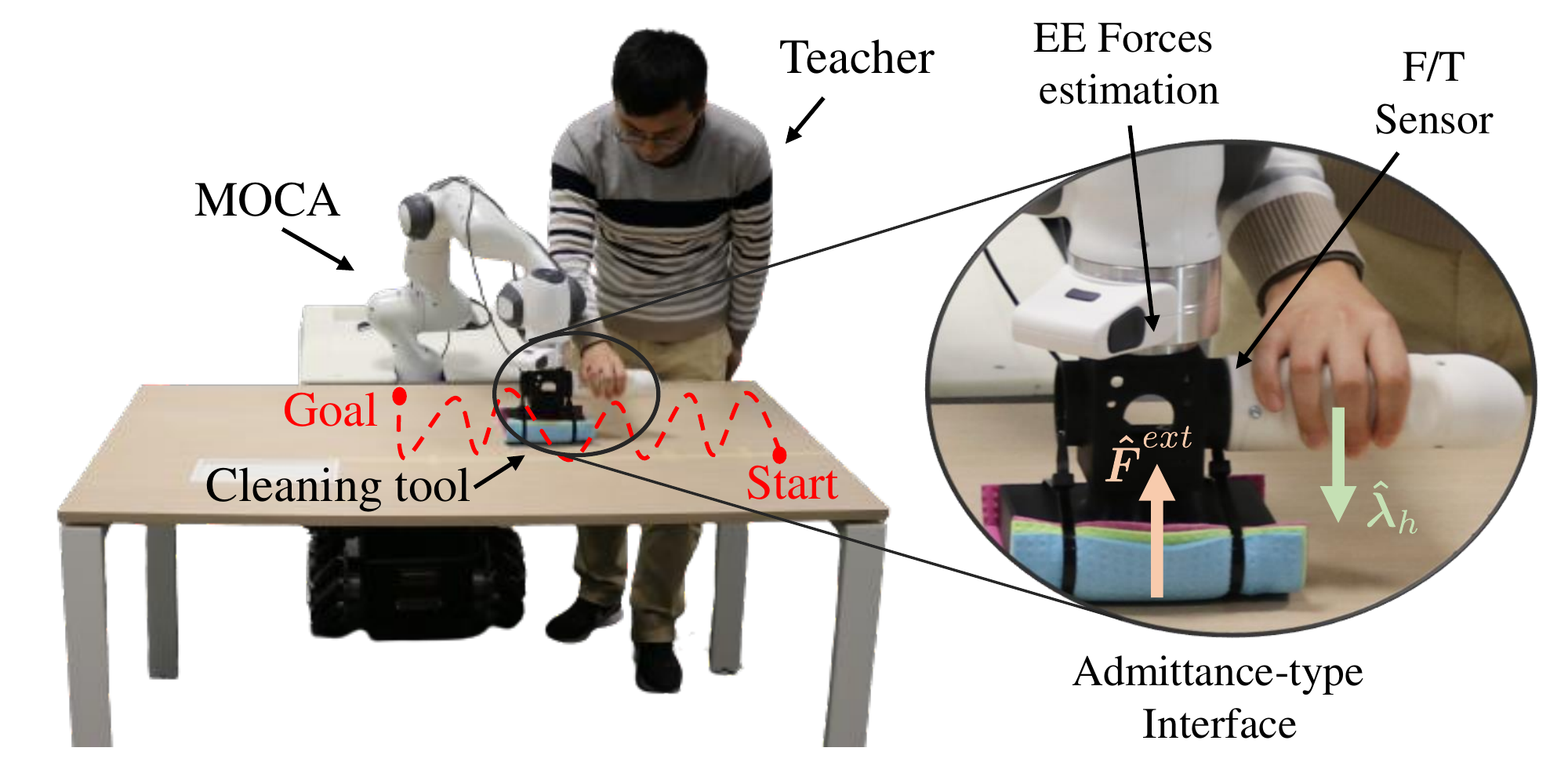}
	\caption{Experimental setup. The teacher demonstrates the table cleaning task to MOCA from Start to Goal. A cleaning tool is attached to the end-effector of MOCA. The end-effector forces $\boldsymbol{\hat{F}}^{ext}$ are estimated thanks to the torque sensors integrated in the robot joints. An additional F/T sensor is included in the admittance-type interface to measure human interaction wrenches $\boldsymbol{\hat{\lambda}}_h$ and decouple them from $\boldsymbol{\hat{F}}^{ext}$.}
	\label{fig:experiment_setup}
	\vspace{-3mm}
\end{figure}

The experiments carried out in this work consist of human demonstrations to train the GMM model and autonomous repetition of a table cleaning task in two different conditions, i.e. without and with external disturbances. For each condition, three stiffness settings are tested: i) low constant stiffness (LS), where $\boldsymbol{k}^d = \boldsymbol{k}^{min}$, ii) high constant stiffness (HS), where $\boldsymbol{k}^d = \boldsymbol{k}^{max}$ and iii) optimized stiffness (OS), where $\boldsymbol{k}^d$ is found online through our QP formulation in \eqref{eq:specific_QP_formulation}. The QP solution was computed in C++ using ALGLIB QP-BLEIC solver\footnote{\url{www.alglib.net/optimization/quadraticprogramming.php}} on Ubuntu 18.04. All experiments were run on a computer with an Intel Core i7-4790S 3.2 GHz $\times$ 8-cores CPU and 16 GB RAM. The same desired end-effector trajectory, generated by GMR, is used for all the stiffness settings, while the desired interaction force is employed only by OS.

The experimental setup for the human demonstrations is shown in Fig.~\ref{fig:experiment_setup} along with the path followed by the human. The human teacher grasps an admittance-type physical interface and can use the four buttons as described in section \ref{subsec:kinesthetic_teaching}. The demonstrated trajectory starts from the \textit{start}, which coincides with the top-left part of the table from the human viewpoint. Then, the human guides the robot end-effector to clean the table, moving parallel to the shortest side of the table, keeping the sponge in contact with the table's surface. When the bottom of the table is reached, the human guides the robot in free motion to the top of the table, shifted to the right w.r.t. the previous trajectory of an amount equal to the cleaning tool width. This procedure is repeated six times until the \textit{goal} is reached. Three demonstrations are performed, all from the same human teacher, where desired position and interaction force of the end-effector with the environment are recorded and are later used to train the GMM. 

\begin{figure*}[t]
    \centering
    \includegraphics[trim=0.0cm 0.0cm 0.0cm 0.0cm,clip,width=0.28\columnwidth]{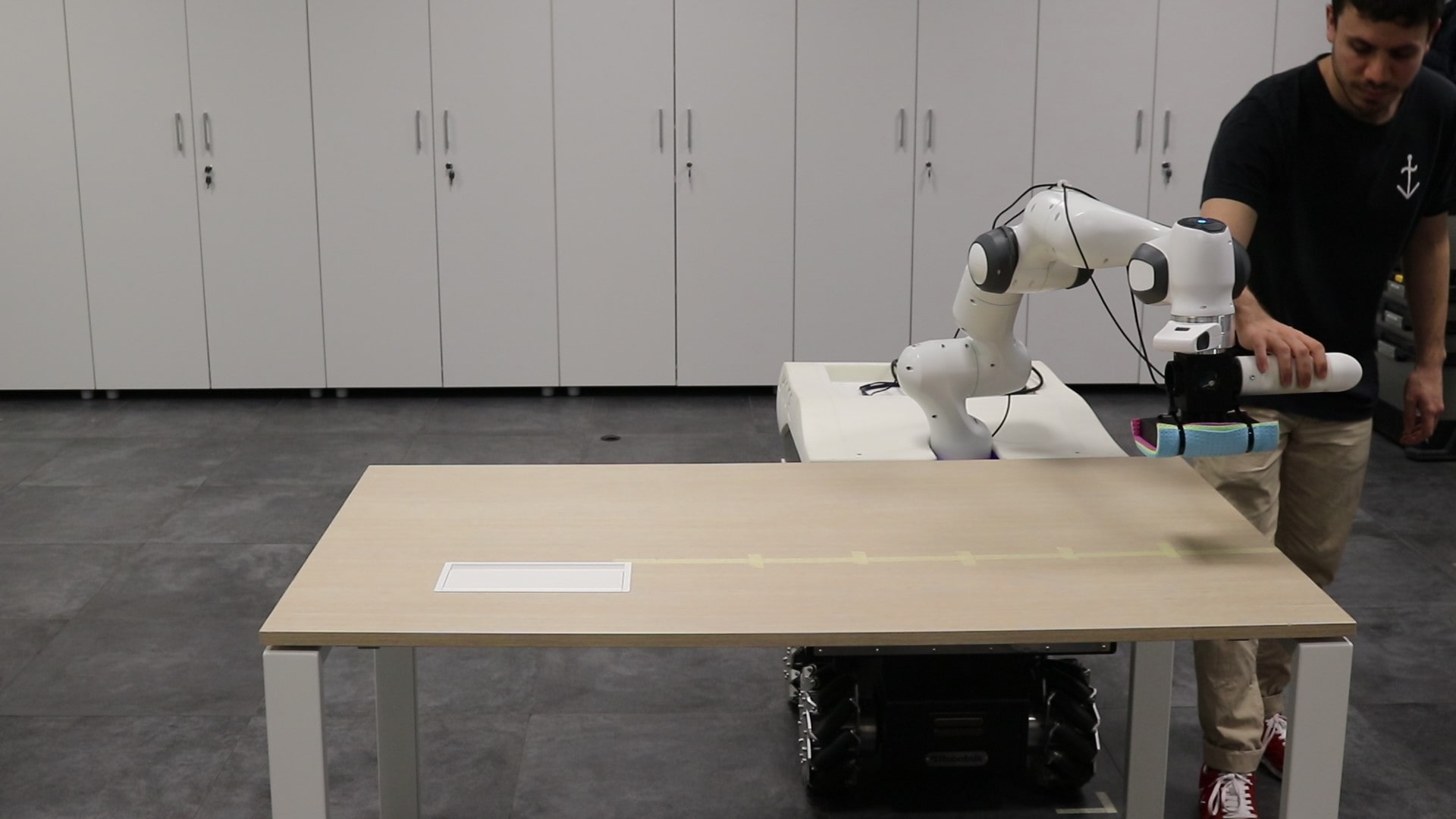}
    \includegraphics[trim=0.0cm 0.0cm 0.0cm 0.0cm,clip,width=0.28\columnwidth]{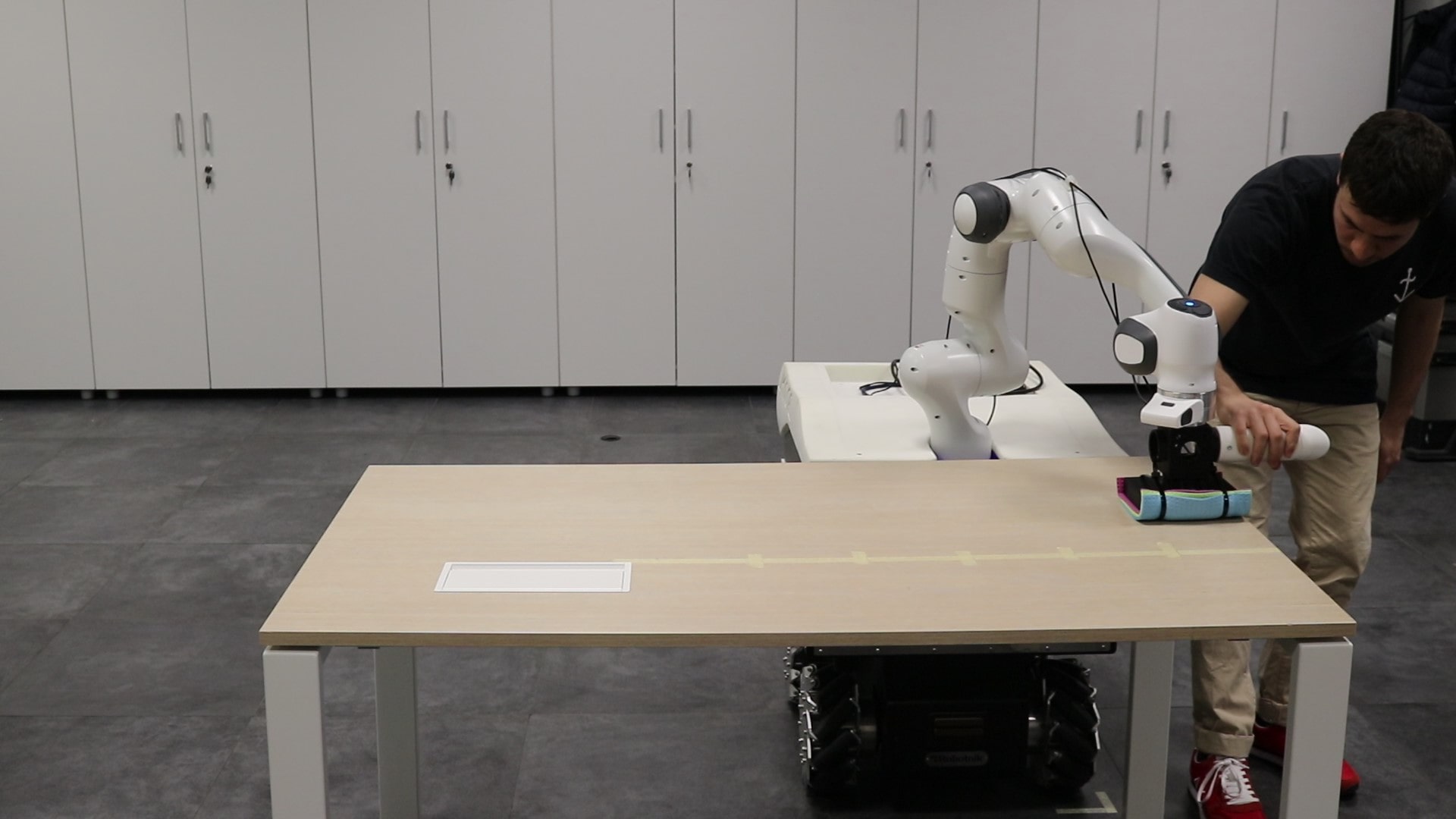}
    \includegraphics[trim=0.0cm 0.0cm 0.0cm 0.0cm,clip,width=0.28\columnwidth]{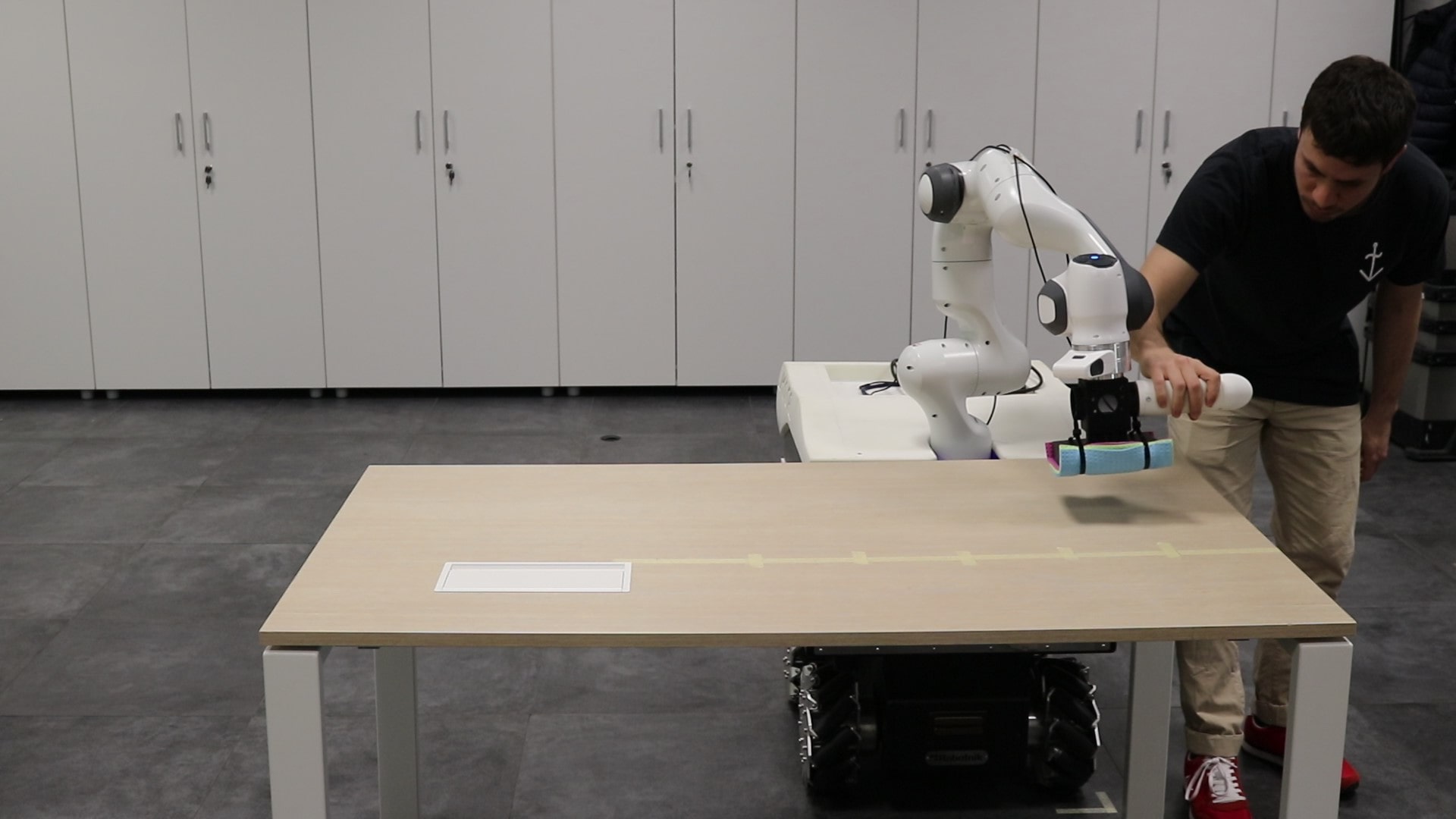}
    \includegraphics[trim=0.0cm 0.0cm 0.0cm 0.0cm,clip,width=0.28\columnwidth]{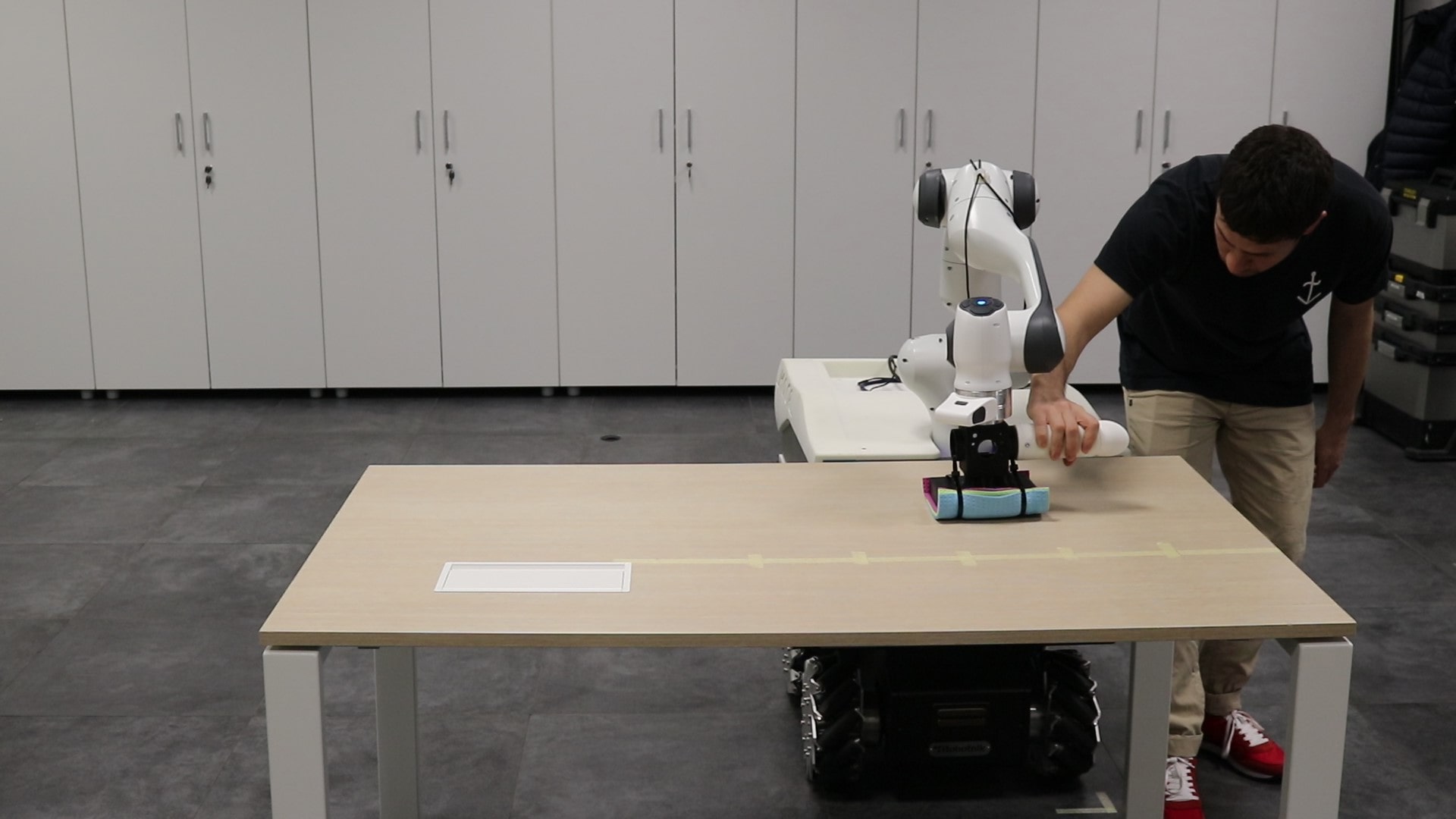}
    \includegraphics[trim=0.0cm 0.0cm 0.0cm 0.0cm,clip,width=0.28\columnwidth]{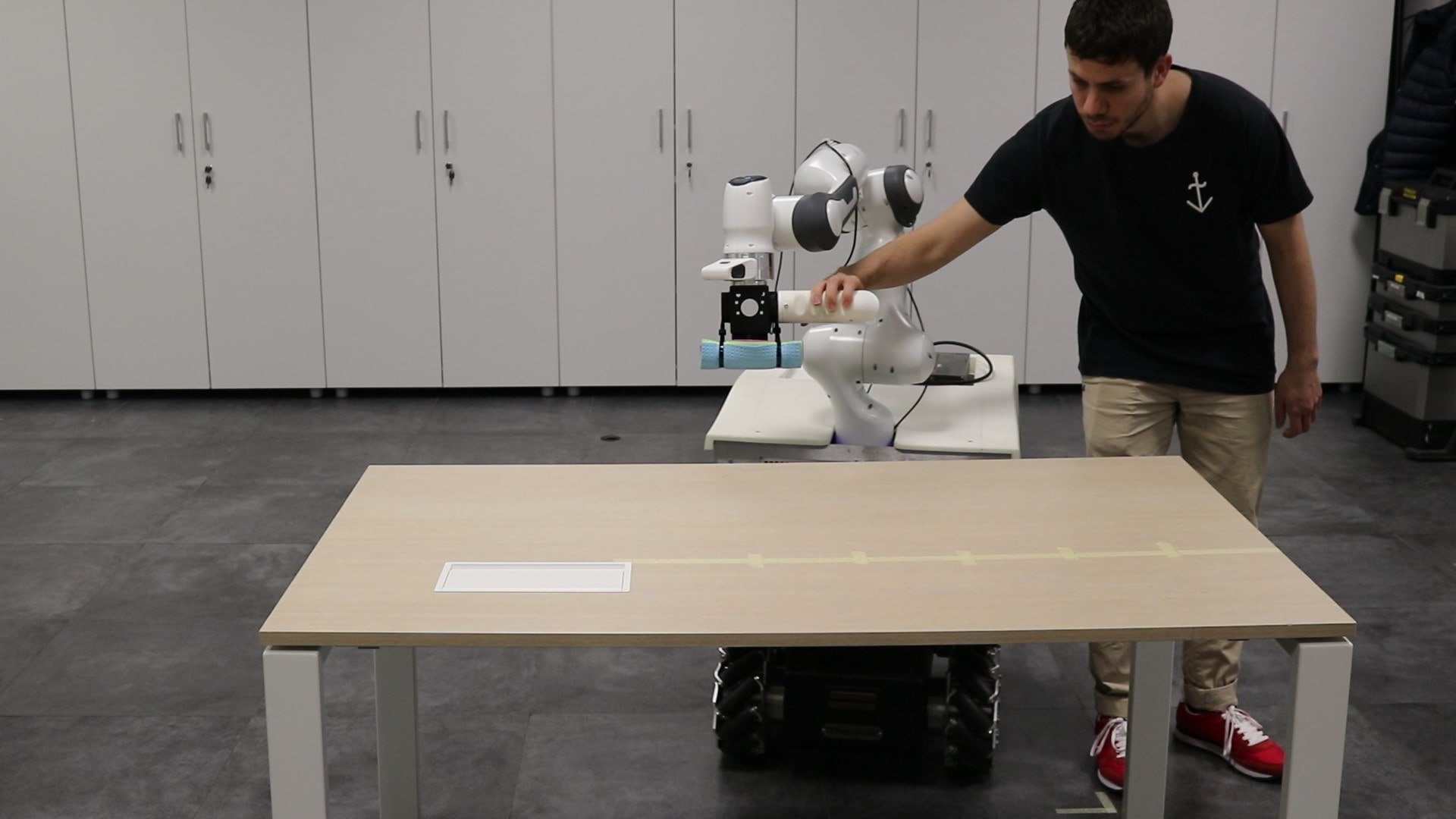}
    \includegraphics[trim=0.0cm 0.0cm 0.0cm 0.0cm,clip,width=0.28\columnwidth]{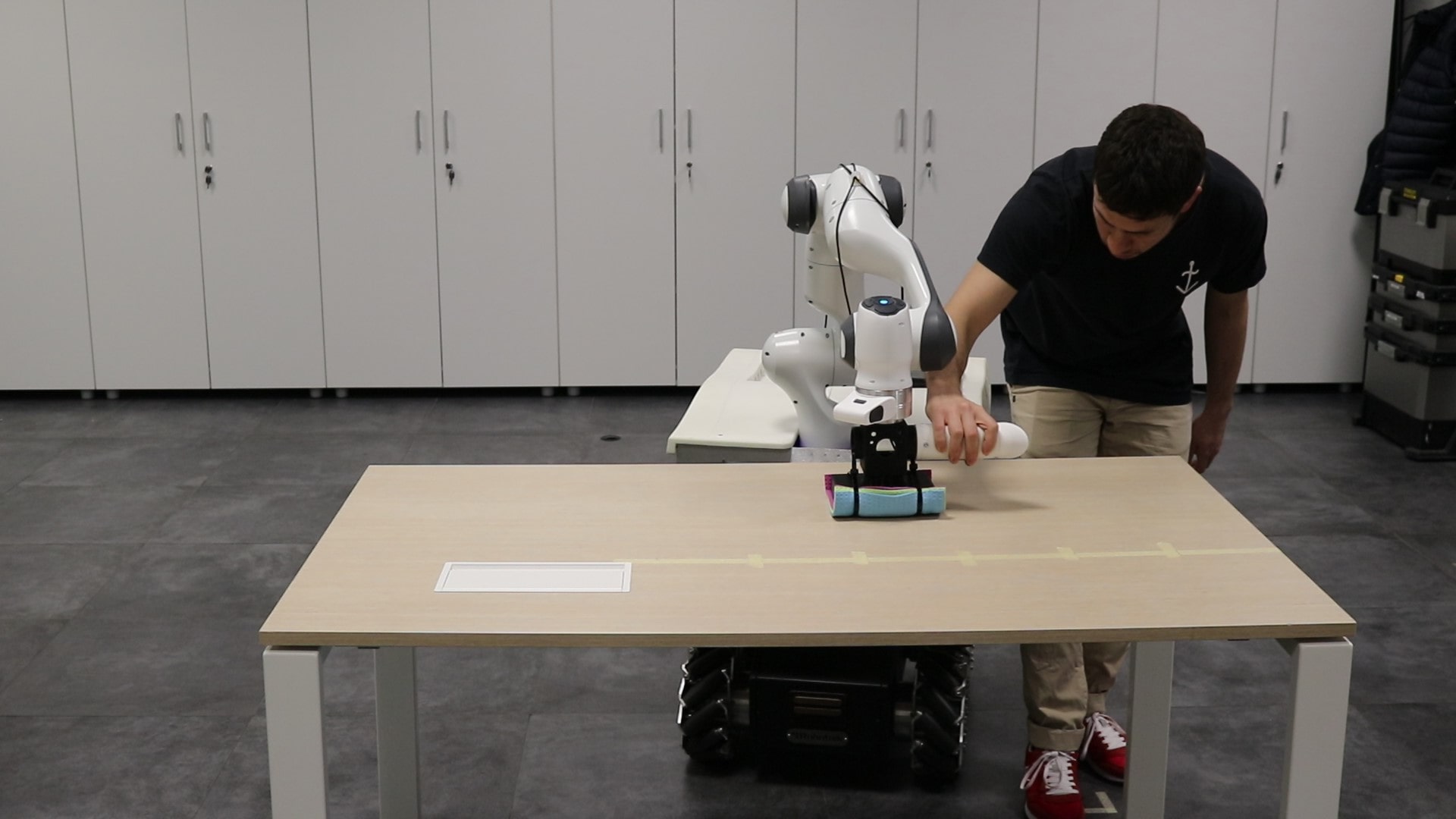}
    \includegraphics[trim=0.0cm 0.0cm 0.0cm 0.0cm,clip,width=0.28\columnwidth]{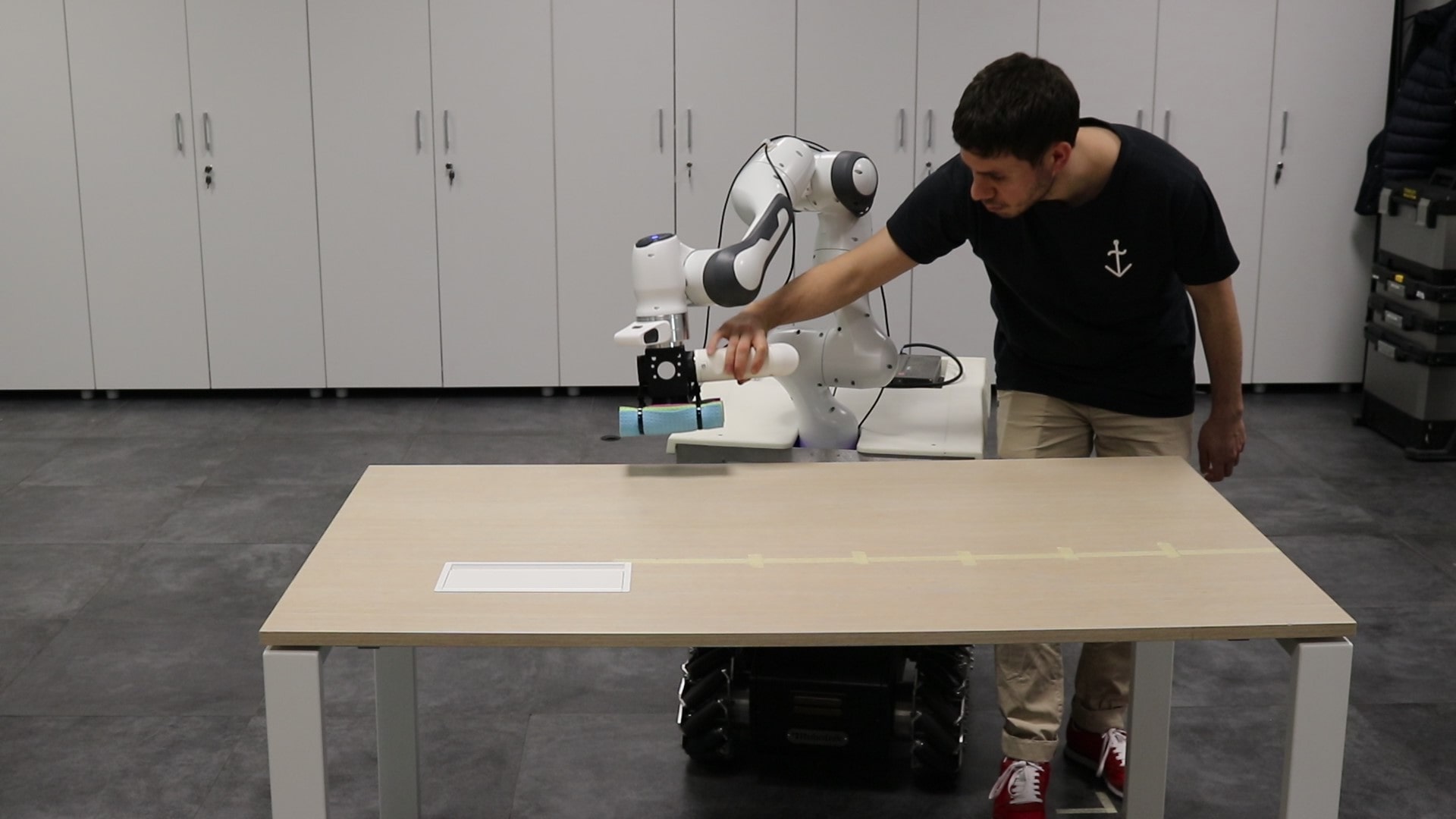}\\
    \vspace{2mm}
    \includegraphics[trim=0.0cm 0.0cm 0.0cm 0.0cm,clip,width=0.28\columnwidth]{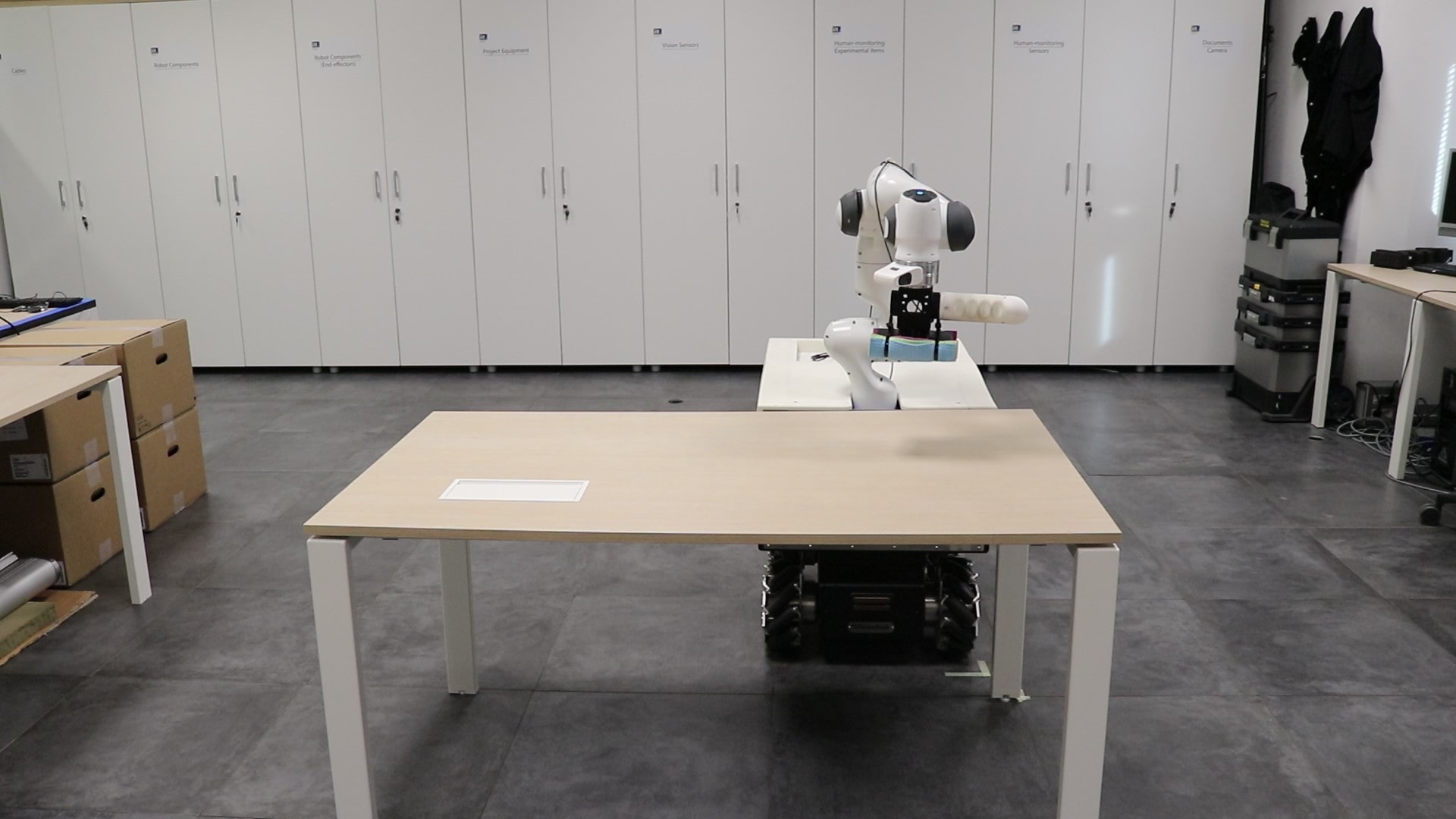}
    \includegraphics[trim=0.0cm 0.0cm 0.0cm 0.0cm,clip,width=0.28\columnwidth]{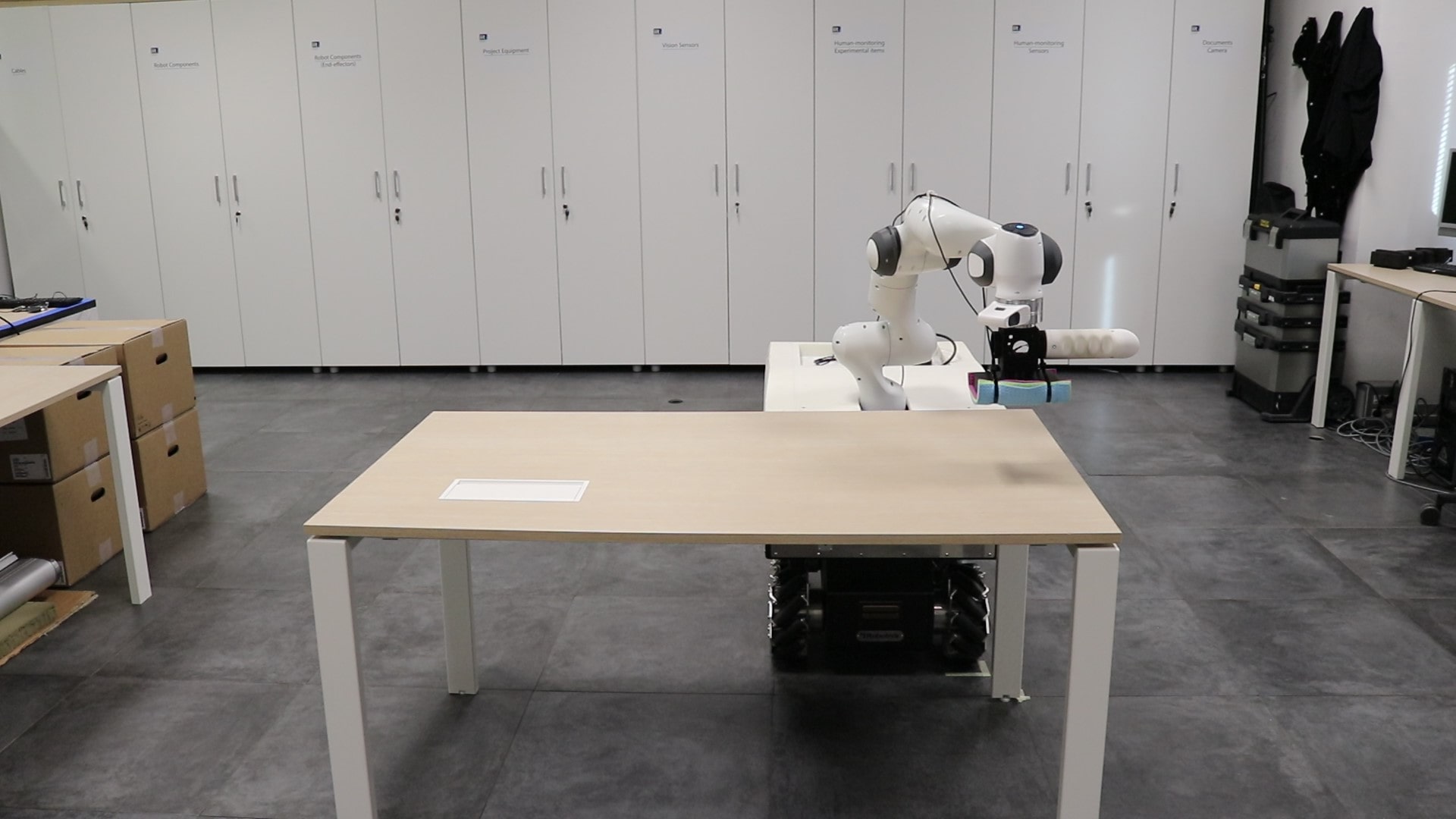}
    \includegraphics[trim=0.0cm 0.0cm 0.0cm 0.0cm,clip,width=0.28\columnwidth]{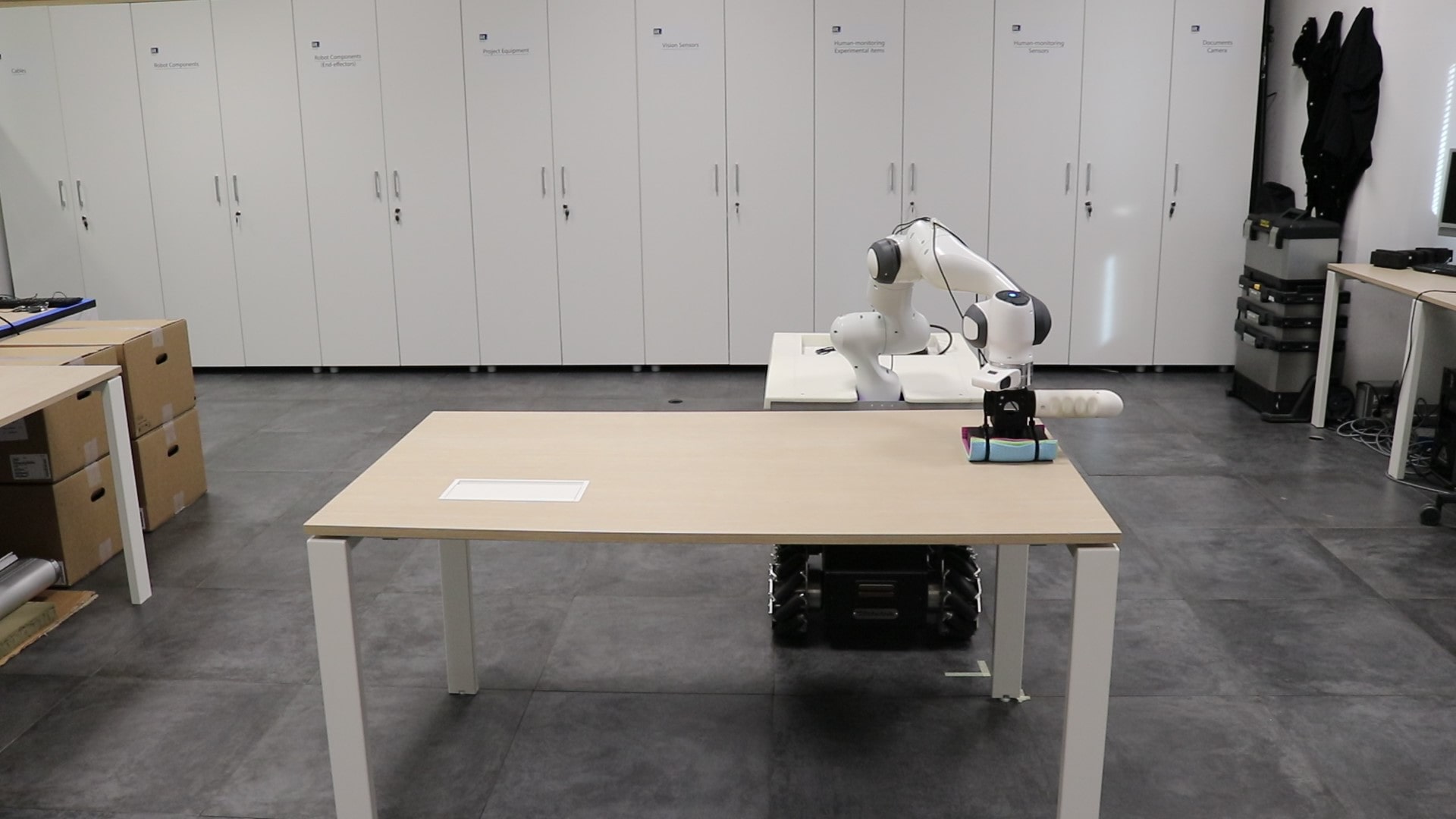}
    \includegraphics[trim=0.0cm 0.0cm 0.0cm 0.0cm,clip,width=0.28\columnwidth]{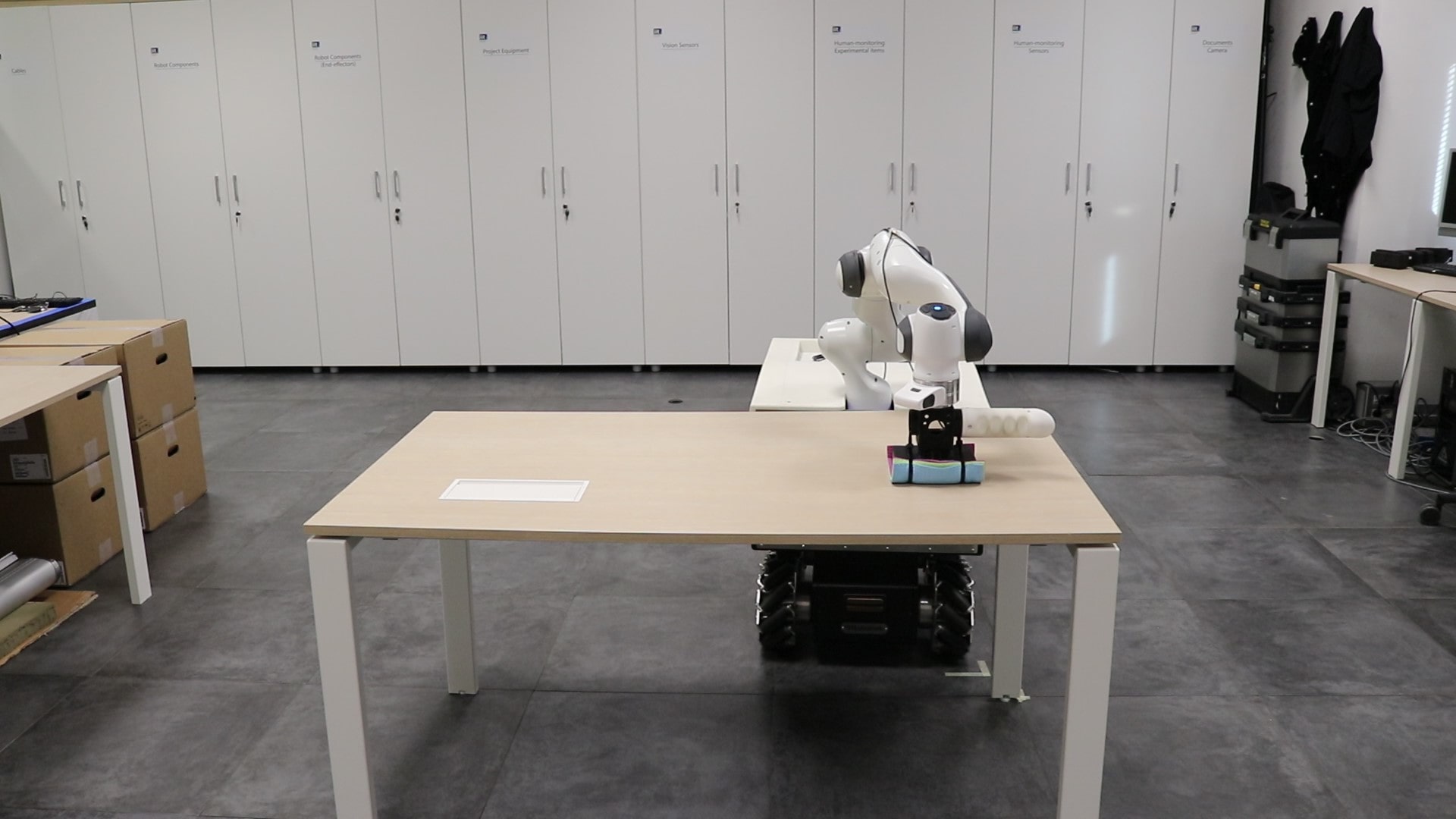}
    \includegraphics[trim=0.0cm 0.0cm 0.0cm 0.0cm,clip,width=0.28\columnwidth]{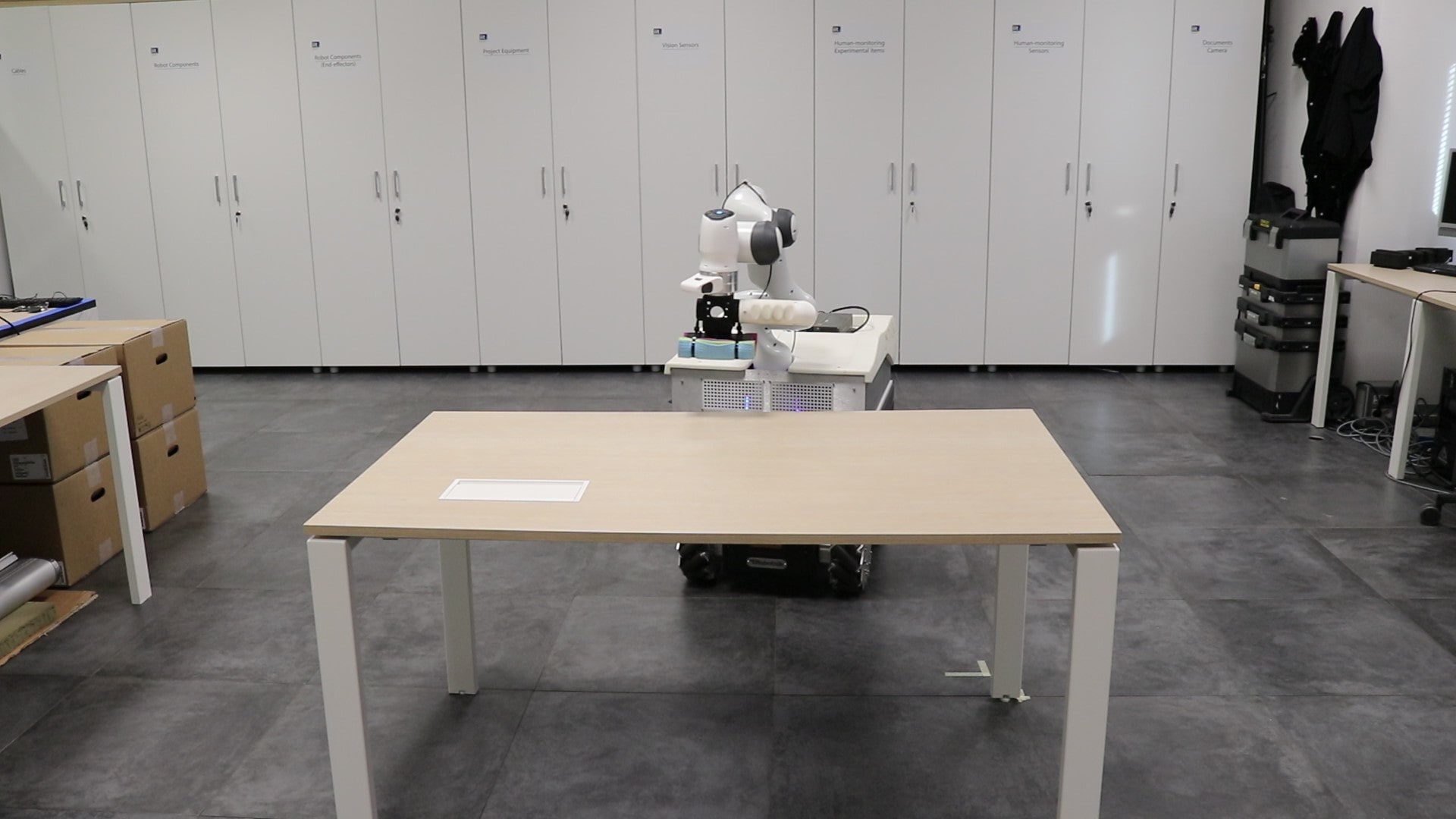}
    \includegraphics[trim=0.0cm 0.0cm 0.0cm 0.0cm,clip,width=0.28\columnwidth]{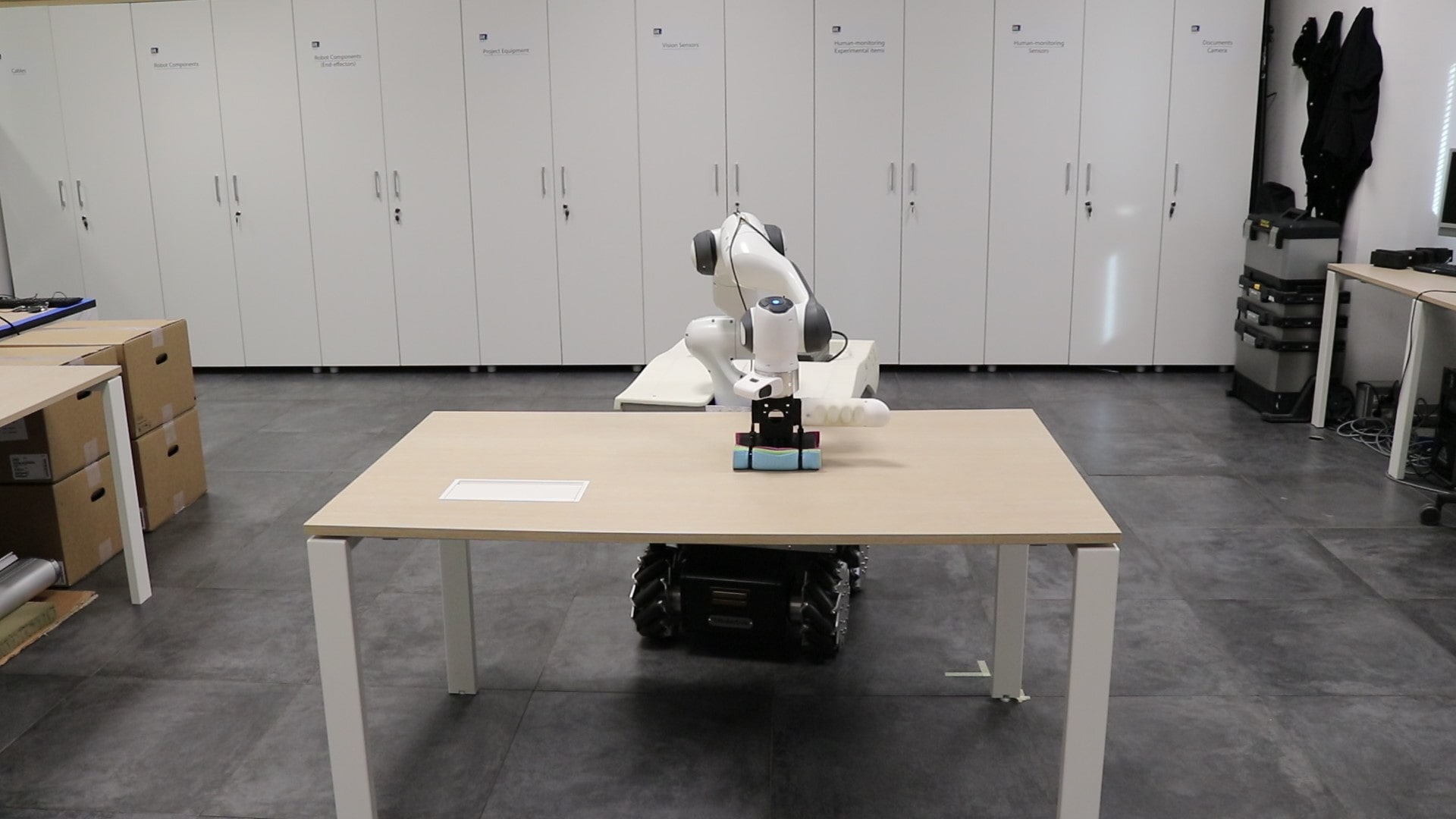}
    \includegraphics[trim=0.0cm 0.0cm 0.0cm 0.0cm,clip,width=0.28\columnwidth]{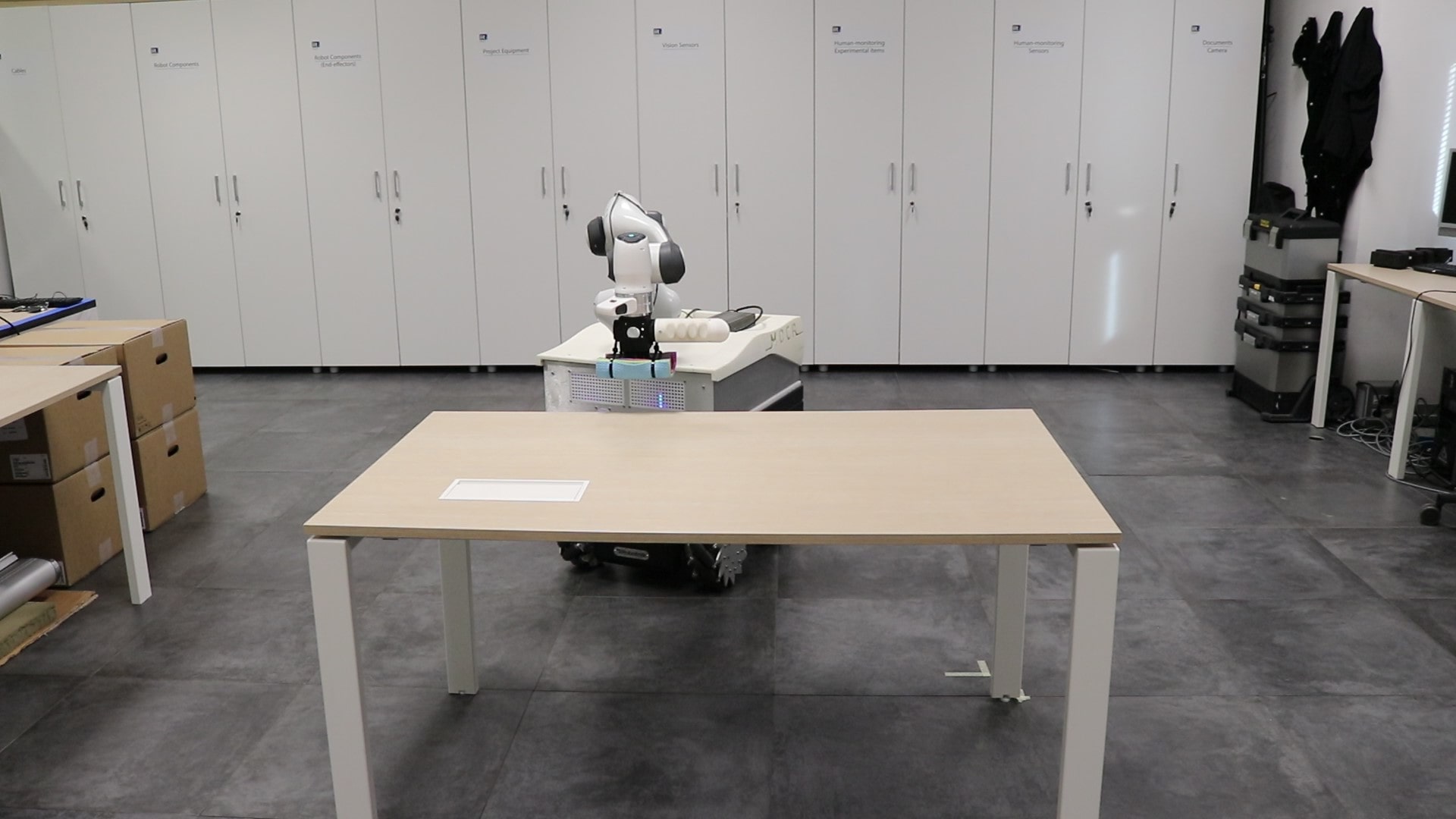}\\
    \vspace{2mm}
    \includegraphics[trim=0.0cm 0.0cm 0.0cm 0.0cm,clip,width=0.28\columnwidth]{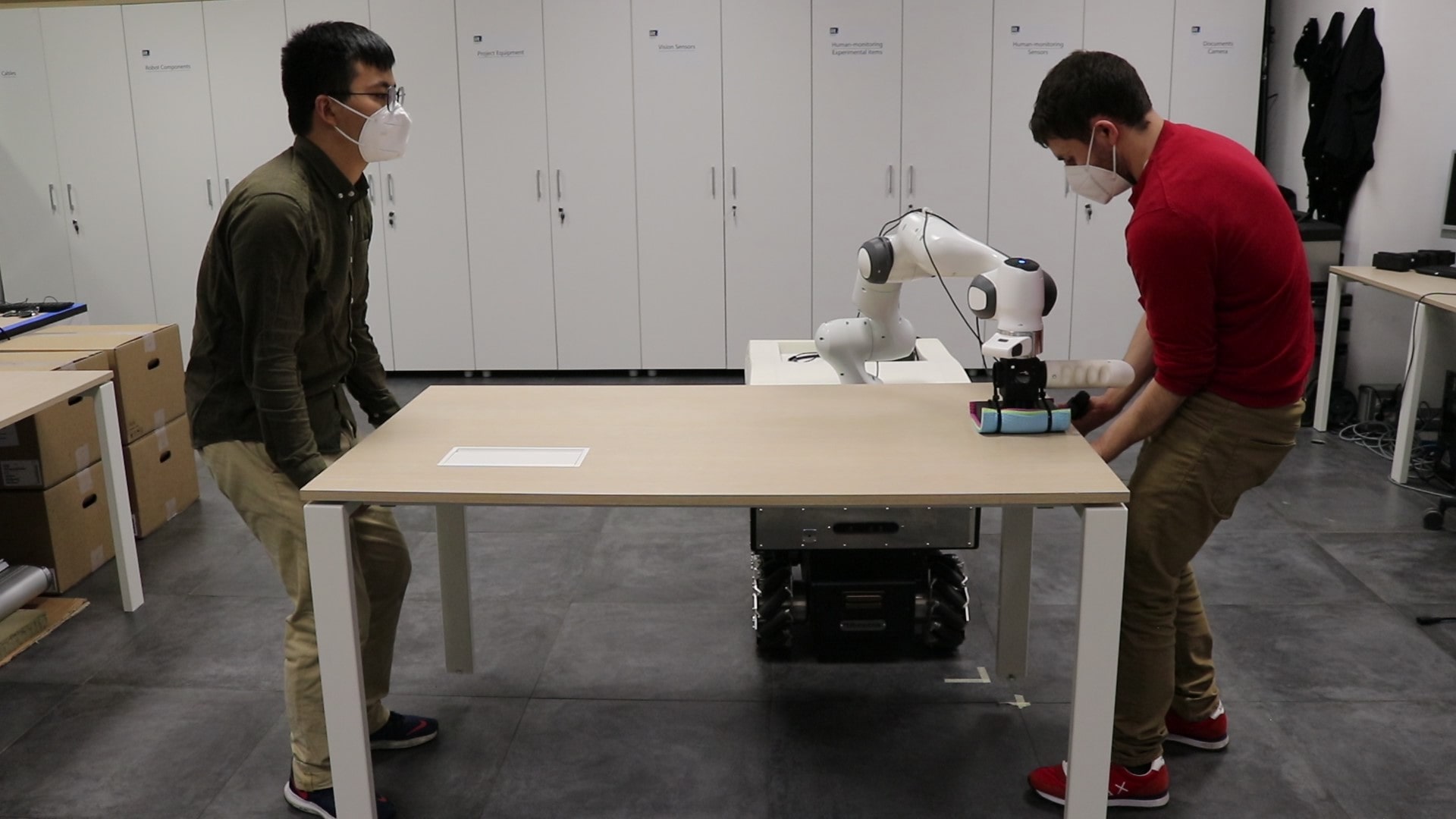}
    \includegraphics[trim=0.0cm 0.0cm 0.0cm 0.0cm,clip,width=0.28\columnwidth]{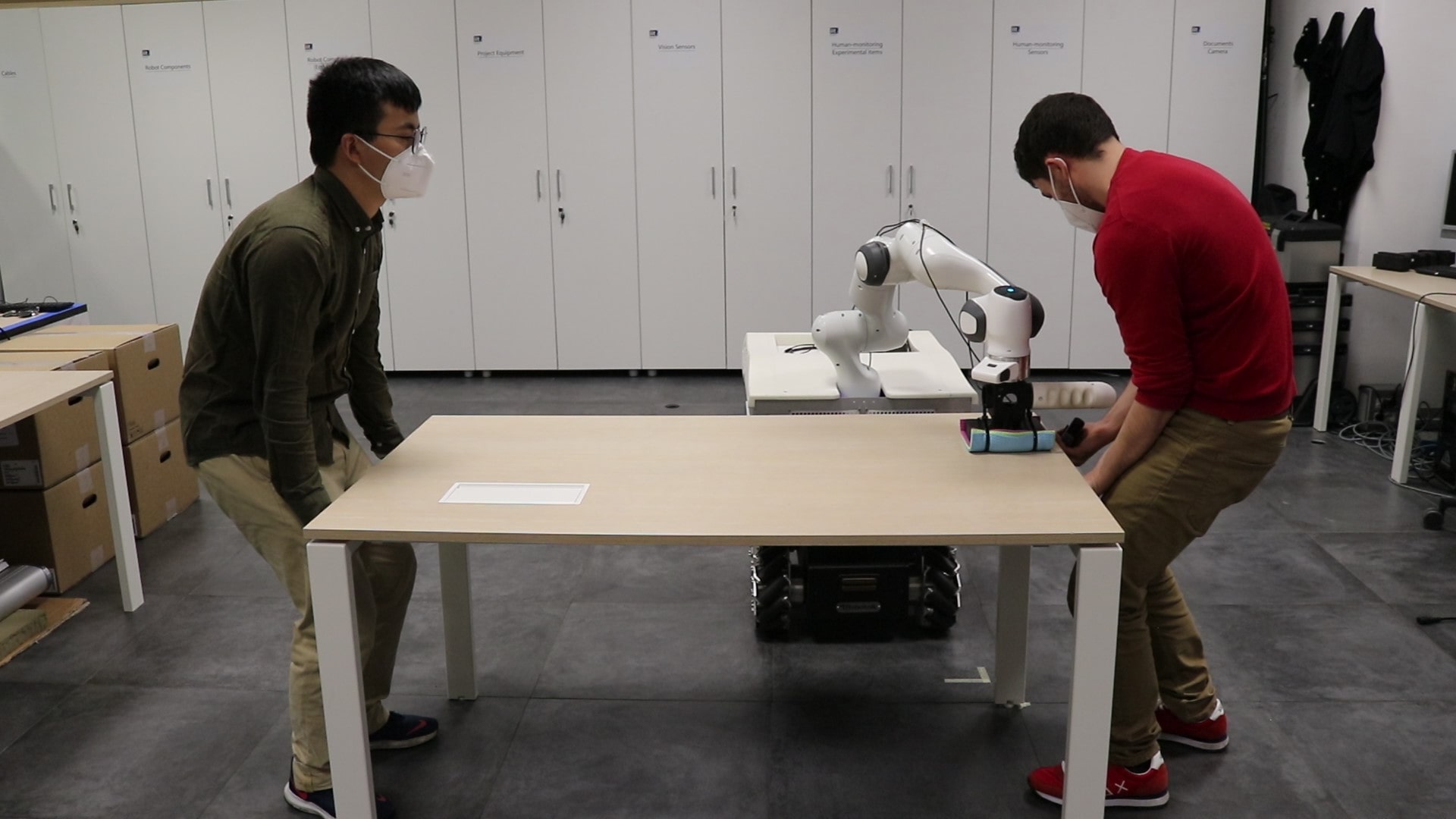}
    \includegraphics[trim=0.0cm 0.0cm 0.0cm 0.0cm,clip,width=0.28\columnwidth]{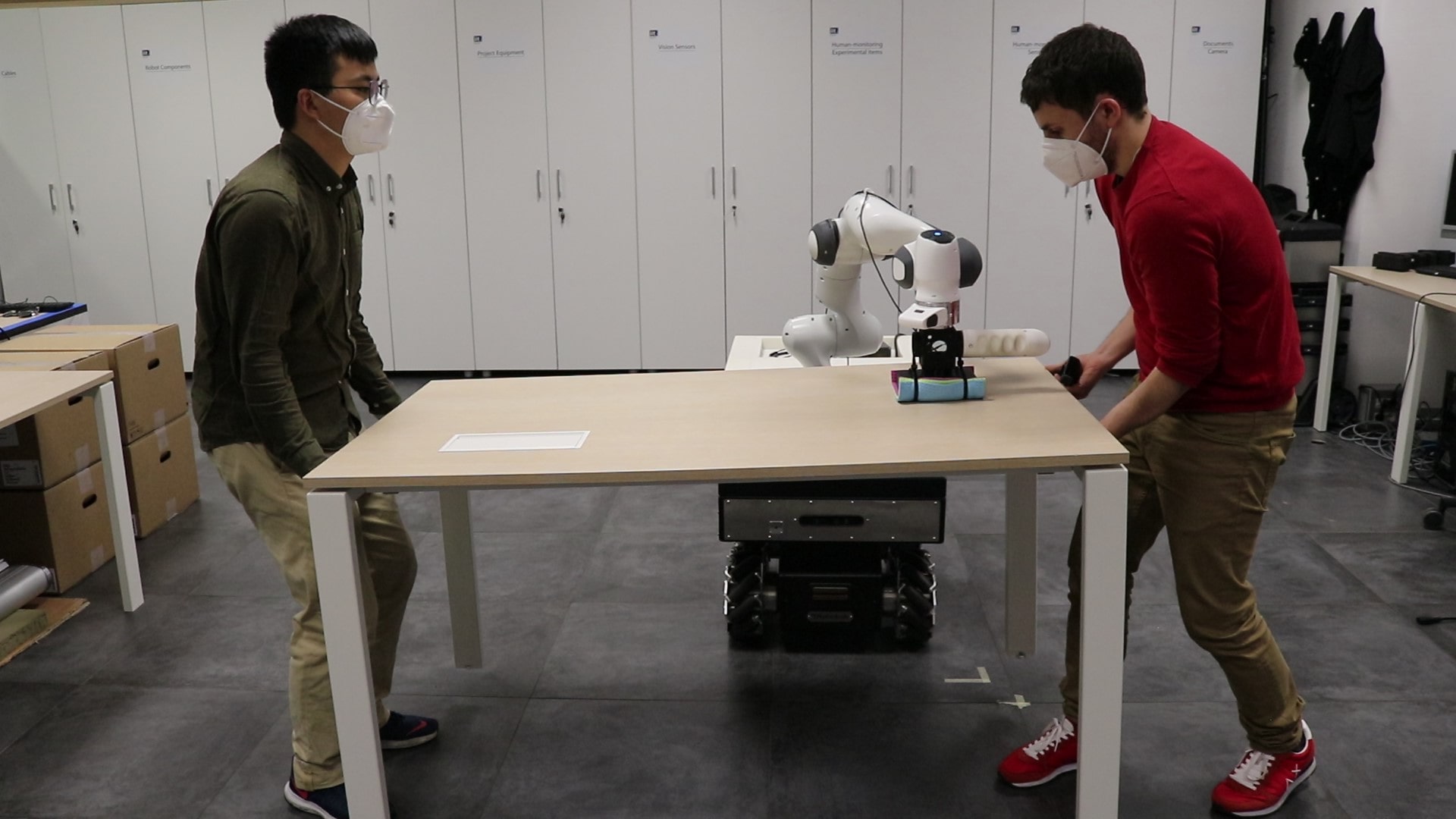}
    \includegraphics[trim=0.0cm 0.0cm 0.0cm 0.0cm,clip,width=0.28\columnwidth]{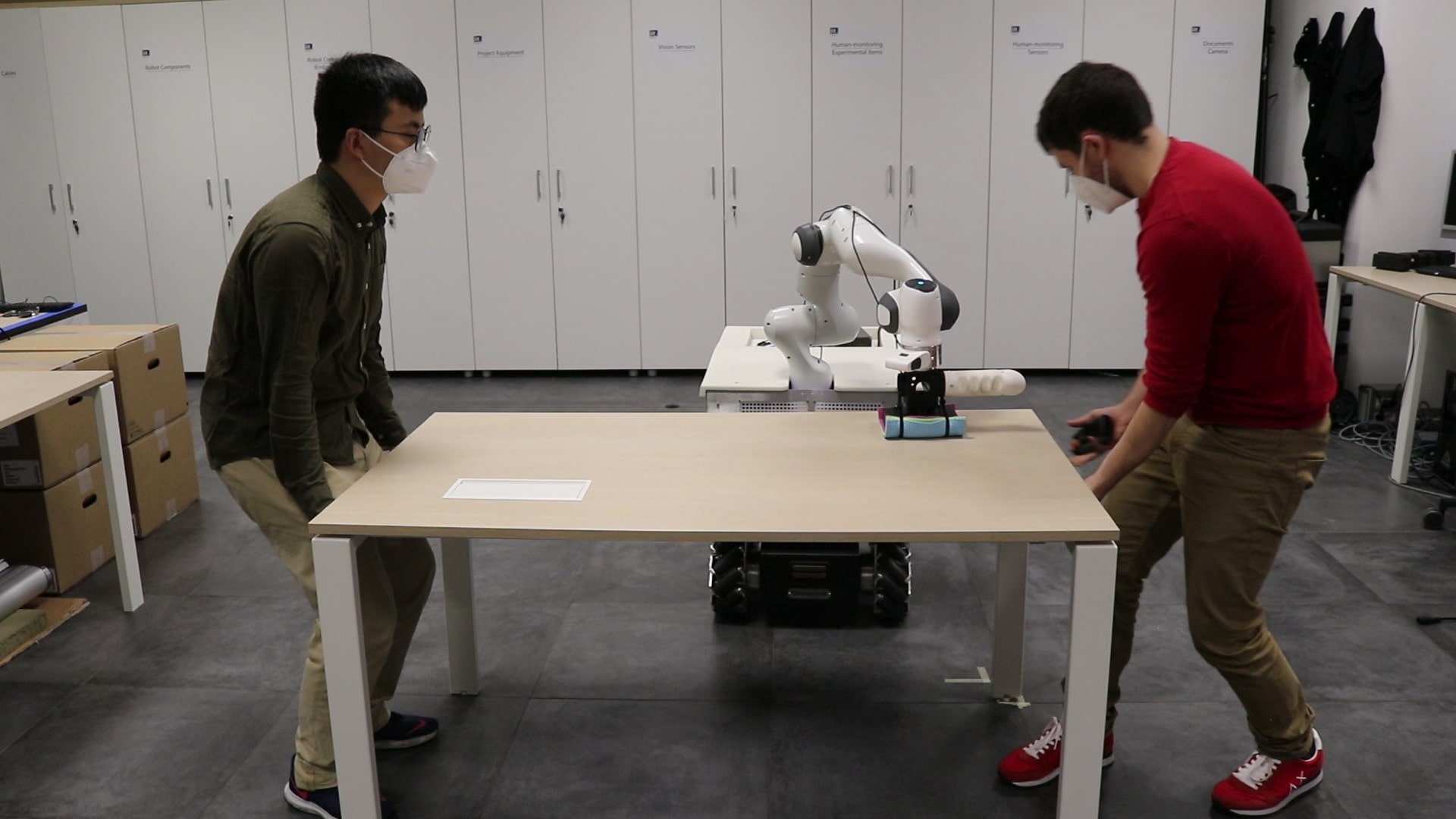}
    \includegraphics[trim=0.0cm 0.0cm 0.0cm 0.0cm,clip,width=0.28\columnwidth]{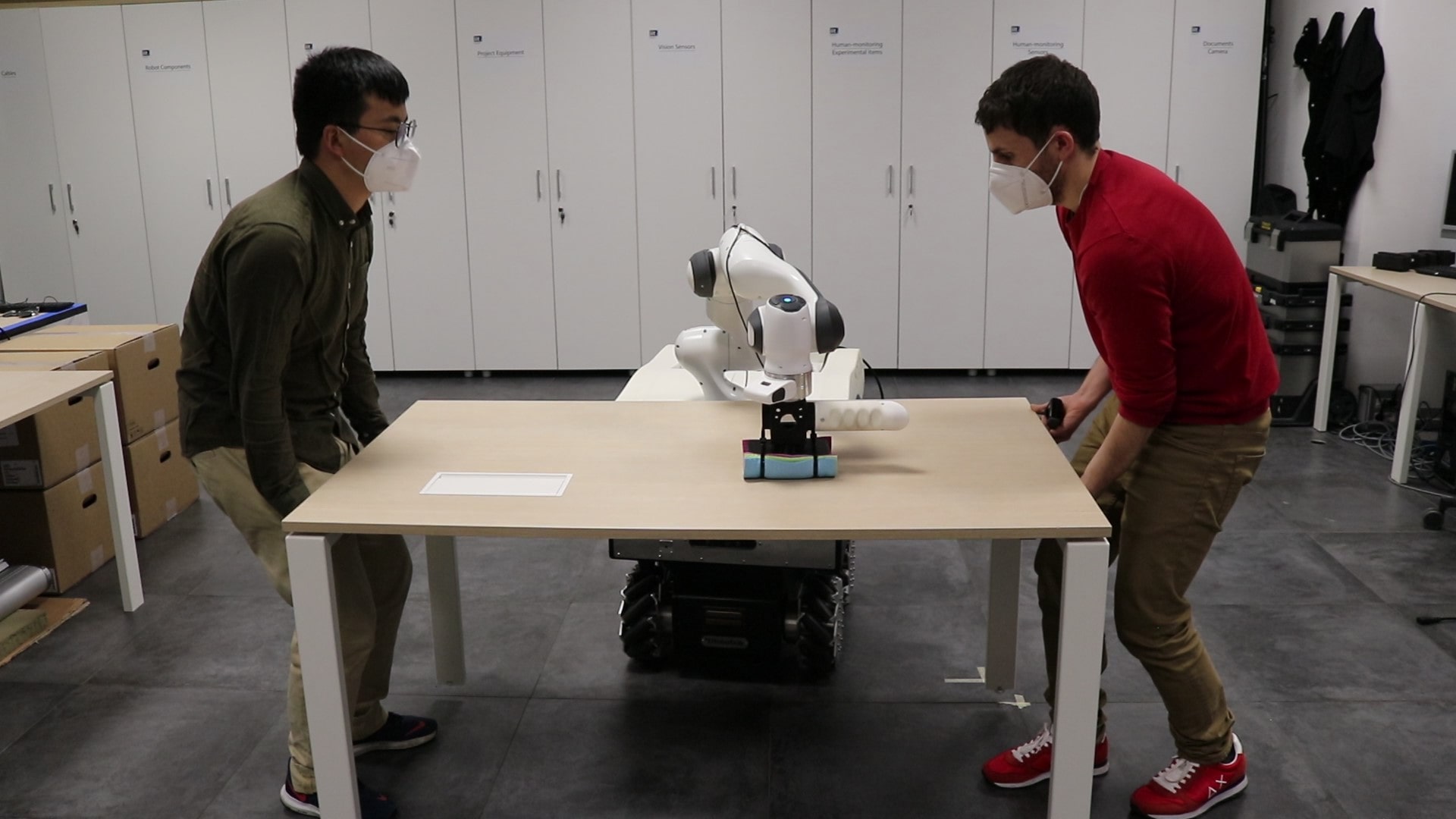}
    \includegraphics[trim=0.0cm 0.0cm 0.0cm 0.0cm,clip,width=0.28\columnwidth]{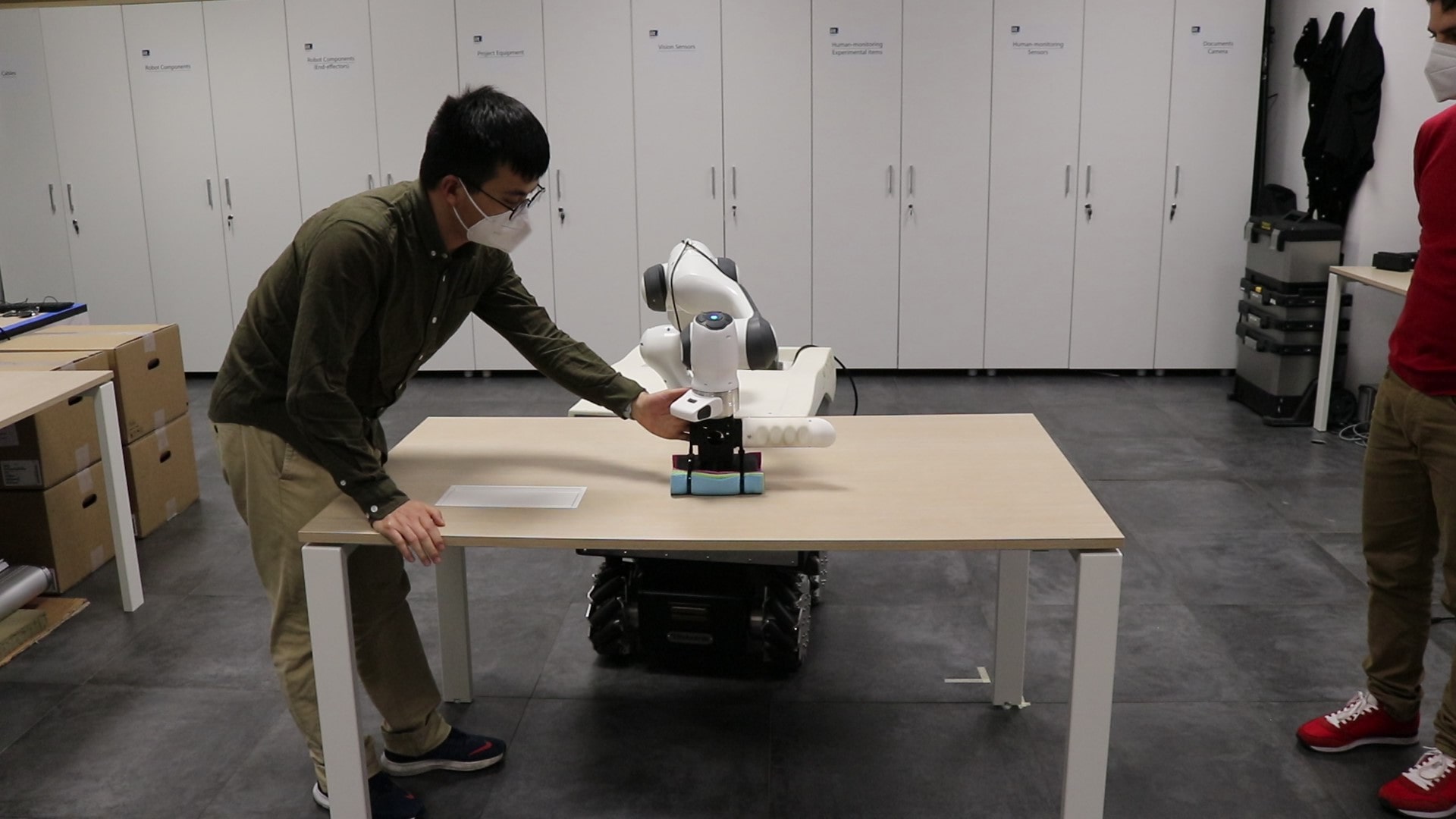}
    \includegraphics[trim=0.0cm 0.0cm 0.0cm 0.0cm,clip,width=0.28\columnwidth]{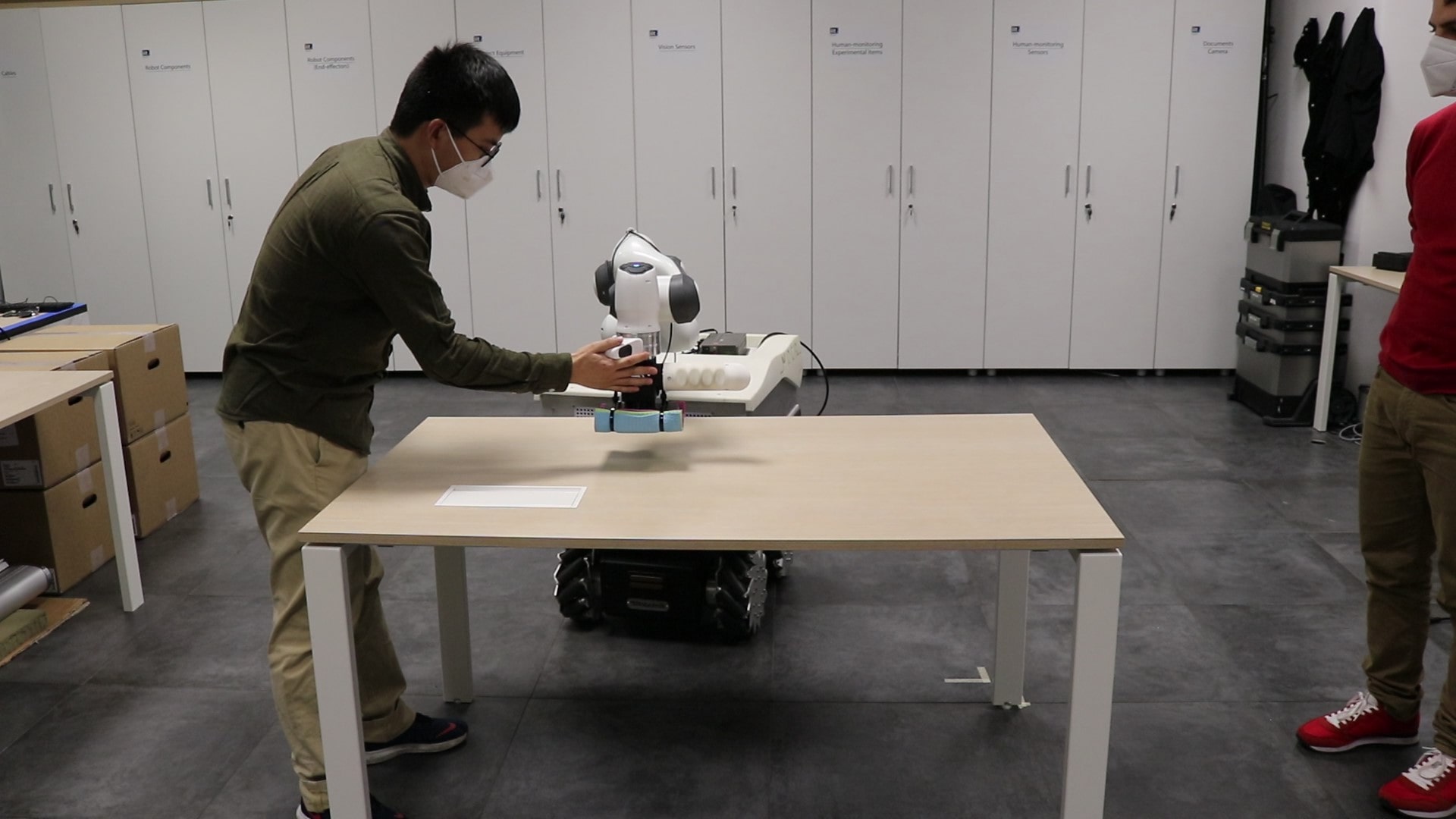}
    \caption{Snapshots of the experiments. (top) Kinesthetic teaching of MOCA using an admittance-type physical interface. (middle) Experiment 1: autonomous task execution in nominal conditions. (bottom) Experiment 2: autonomous task execution with random disturbances. The desk's height was changed in different frequencies and scales in the first five photos. The human stopped MOCA's end-effector during the cleaning and free motion phase in the last two photos, respectively. A video of the experiments is available in the multimedia extension and at the link: \url{https://youtu.be/wotzfcpcAU0}.}
    \label{fig:snapshots}
    \vspace{-3mm}
\end{figure*}

A constant Cartesian impedance for the whole-body controller of MOCA is used during the demonstrations, with a stiffness of $\boldsymbol{K}^d$=$diag\{500,500,500,50,50,50\}$ and a damping computed through the double diagonalization formula. Also, joint stiffness is added  as secondary task to prevent joint limits reaching and low manipulability configurations, $\boldsymbol{\tau}_{0} = -\boldsymbol{D}_0\dot{\boldsymbol{q}} - \boldsymbol{K}_0(\boldsymbol{q} - \boldsymbol{q}_0)$, where $\boldsymbol{K}_0, \boldsymbol{D}_0 \in \mathbb{R}^{n \times n}$ are joint stiffness and damping, while $ \boldsymbol{q}_0 \in \mathbb{R}^n$ is the desired joint configuration. The teacher can change the admittance level to three discrete values corresponding to low ($\boldsymbol{M}^{adm}$=$diag\{6,6,6\}$, $\boldsymbol{D}^{adm}$=$diag\{40,40,40\}$), medium ($\boldsymbol{M}^{adm}$=$diag\{4,4,4\}$, $\boldsymbol{D}^{adm}$=$diag\{30,30,30\}$) and high ($\boldsymbol{M}^{adm}$=$diag\{2,2,2\}$, $\boldsymbol{D}^{adm}$=$diag\{20,20,20\}$) admittance using a button of an admittance-type physical interface. However, a constant high admittance is always chosen, in order to make the human forces negligible w.r.t. the ones coming from the environment. Moreover, the teacher can also switch between locomotion (large mobile base movements) and manipulation (mobile base fixed) mode, since a button of an admittance-type physical interface allows to change $\boldsymbol{H}$, $\boldsymbol{K}_0$ and $\boldsymbol{q}_0$ simultaneously. Particularly, $\boldsymbol{H}$ and $\boldsymbol{K}_0$ can take on two discrete values ($\boldsymbol{H}$=$diag\{10 \cdot \boldsymbol{1}_{n_b},2\cdot\boldsymbol{1}_{n_a}\}$, $\boldsymbol{K}_0$=$diag\{2\cdot\boldsymbol{1}_n\}$ for manipulation and $\boldsymbol{H}$=$diag\{2\cdot\boldsymbol{1}_{n_b},10\cdot\boldsymbol{1}_{n_a}\}$, $\boldsymbol{K}_0$=$diag\{50\cdot\boldsymbol{1}_n\}$ for locomotion), while $\boldsymbol{q}_0$ is set to the current arm configuration when a switch between the two modes occurs.

The same experimental setup of the demonstrations is used for the autonomous repetition of the task. Two conditions are tested for each stiffness setting (LS, HS, and OS). In the first one, i.e., the autonomous task repetition without external disturbances, the robot repeats the demonstration in the same environment displayed during the demonstration. While in the second one, i.e., the autonomous task repetition with external disturbances, the environment is perturbed during the execution of the task. As previously mentioned, the cleaning task consists of 6 cleaning movements that go from the top to the bottom of the table, alternated with five free motions to return to the top of the table before starting the next cleaning movement. The following perturbations are applied during the experiment (see also Fig.~\ref{fig:snapshots} (bottom)):
% \vspace{-2mm}
\begin{enumerate}
    \item Single slow table lifting and lowering during the $1^{st}$ cleaning movement.
    \item Single fast table lifting and lowering during the $2^{nd}$ cleaning movement.
    \item Repeated low frequency and high amplitude table lifting and lowering during the $3^{rd}$ cleaning movement.
    \item Repeated high frequency and low amplitude table lifting and lowering during the $4^{th}$ cleaning movement.
    \item Collision with a human during the $5^{th}$ cleaning motion.
    \item Collision with a human during the last free motion.
\end{enumerate}
% \vspace{-2mm}

The values of the parameters used during the autonomous repetition experiments are $\boldsymbol{k}^{min}$=$[200,200,200]$, $\boldsymbol{k}^{max}$=$[1000,1000,1000]$, $\boldsymbol{F}^{max}$=$[60,60,60]$, $\varepsilon$=$0.4$, $\textrm{x}_t(0)$=$1$, $\boldsymbol{Q}$=$diag\{3200,3200,3200\}$ and $\boldsymbol{R}$=$diag\{1,1,1\}$. The minimum stiffness is chosen in order to have high compliance but ensuring an acceptable position tracking performance, while the maximum stiffness is selected in order to track the desired motion with high accuracy while ensuring a stable behavior. The maximum force corresponds to the maximum payload of the robot, and the initial energy of the tank is slightly higher than the energy threshold ($\varepsilon$). Finally, $\boldsymbol{Q}$ and $\boldsymbol{R}$ are chosen experimentally to give higher priority to the force tracking.

\subsection{Experimental Results}

\begin{figure*}[t]
	\centering
	\includegraphics[width=0.95\textwidth,height=7.5cm]{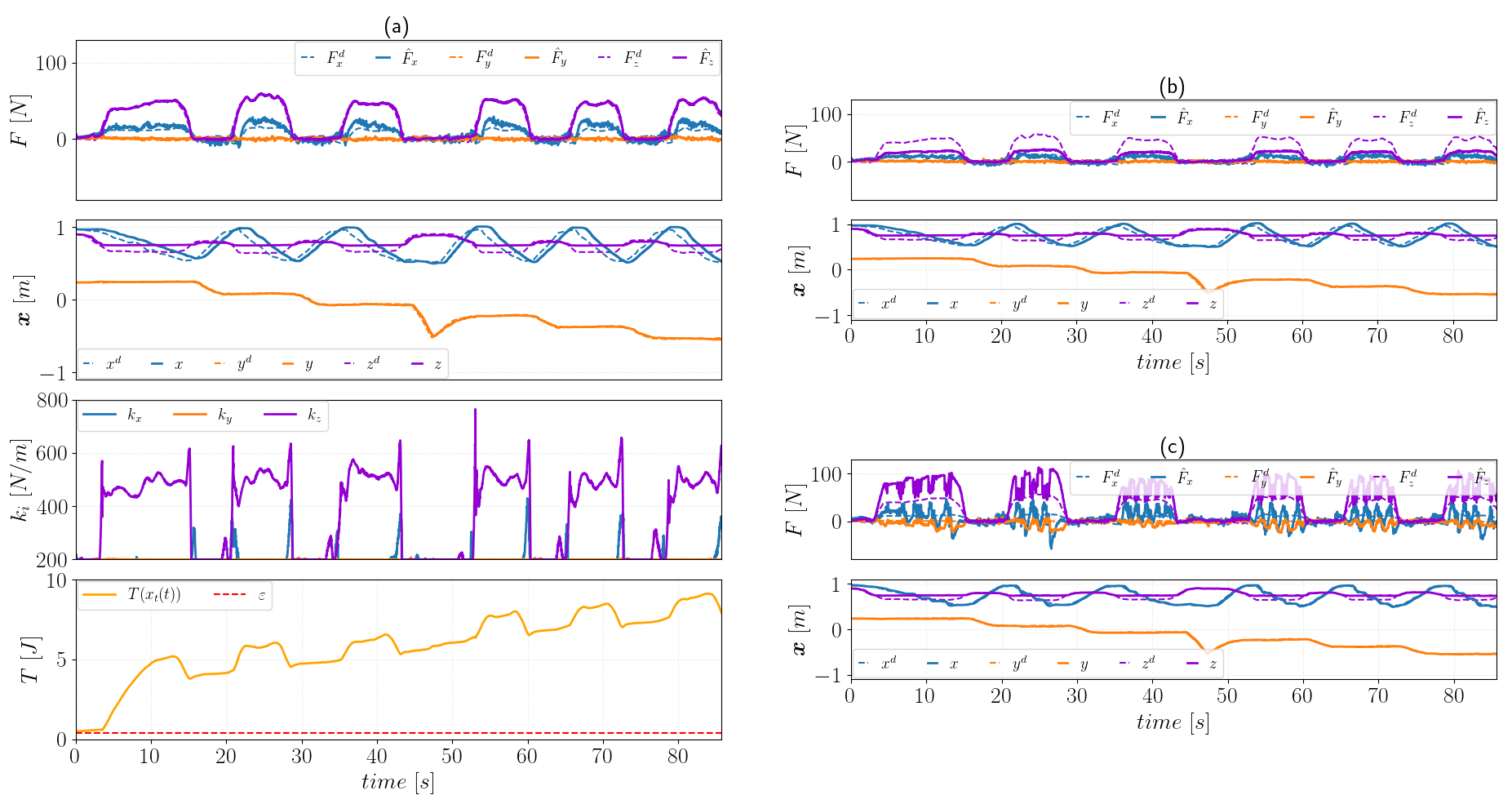}
	\caption{Results of autonomous task execution experiments in nominal conditions. (a). OS settings, from top to bottom are desired interactive force $\boldsymbol{F}^{d}$ and estimated force $\boldsymbol{\hat{F}}^{ext}$ during experiments, desired position $\boldsymbol{X}^{d}$ and real position $\boldsymbol{X}$, generated stiffness $\boldsymbol{K}$, and tank energy $T(\textrm{x}_t)$. (b). LS settings,from top to bottom are desired interactive force $\boldsymbol{F}^{d}$ and estimated force $\boldsymbol{\hat{F}}^{ext}$ during experiments, desired position $\boldsymbol{X}^{d}$ and real position $\boldsymbol{X}$. (c). HS settings, same contents with (b). }  
	\label{fig:repeat_experiment}
	\vspace{-3mm}
\end{figure*}
\label{subsec:results_autonom_exp}

The results of the autonomous task repetition without external disturbances for LS, HS, and OS are shown in Fig.~\ref{fig:repeat_experiment}. For all the experiments, force tracking and position tracking performance are reported, while the stiffness values and energy in the tank are shown only for OS. Although LS allows a compliant behavior, it is not able to track the desired force along $z$, which is necessary for the cleaning performance (Fig.~\ref{fig:repeat_experiment}(b)). Conversely, for HS (Fig.~\ref{fig:repeat_experiment}(c)) a too large force is applied along $z$. This results in the stick and slip phenomenon, as can be viewed from the measured force and position trend along the $x$ axis.
On the other hand, OS (Fig.~\ref{fig:repeat_experiment}(a)) generates the right amount of compliance needed to track the desired force during the interaction with the table while keeping low stiffness in free motion when there are no interactions. The two sources of energy variations in the tank are the dissipated energy and the exchanged energy due to the variable stiffness (see \eqref{eq:Tank_derivative}). The energy in the tank has an increasing trend, and it never falls below the threshold $\varepsilon$, meaning that the passivity of the system is preserved, although sometimes energy drops due to stiffness variations.

\begin{figure*}[t]
	\centering
	\includegraphics[width=\textwidth,height=8cm]{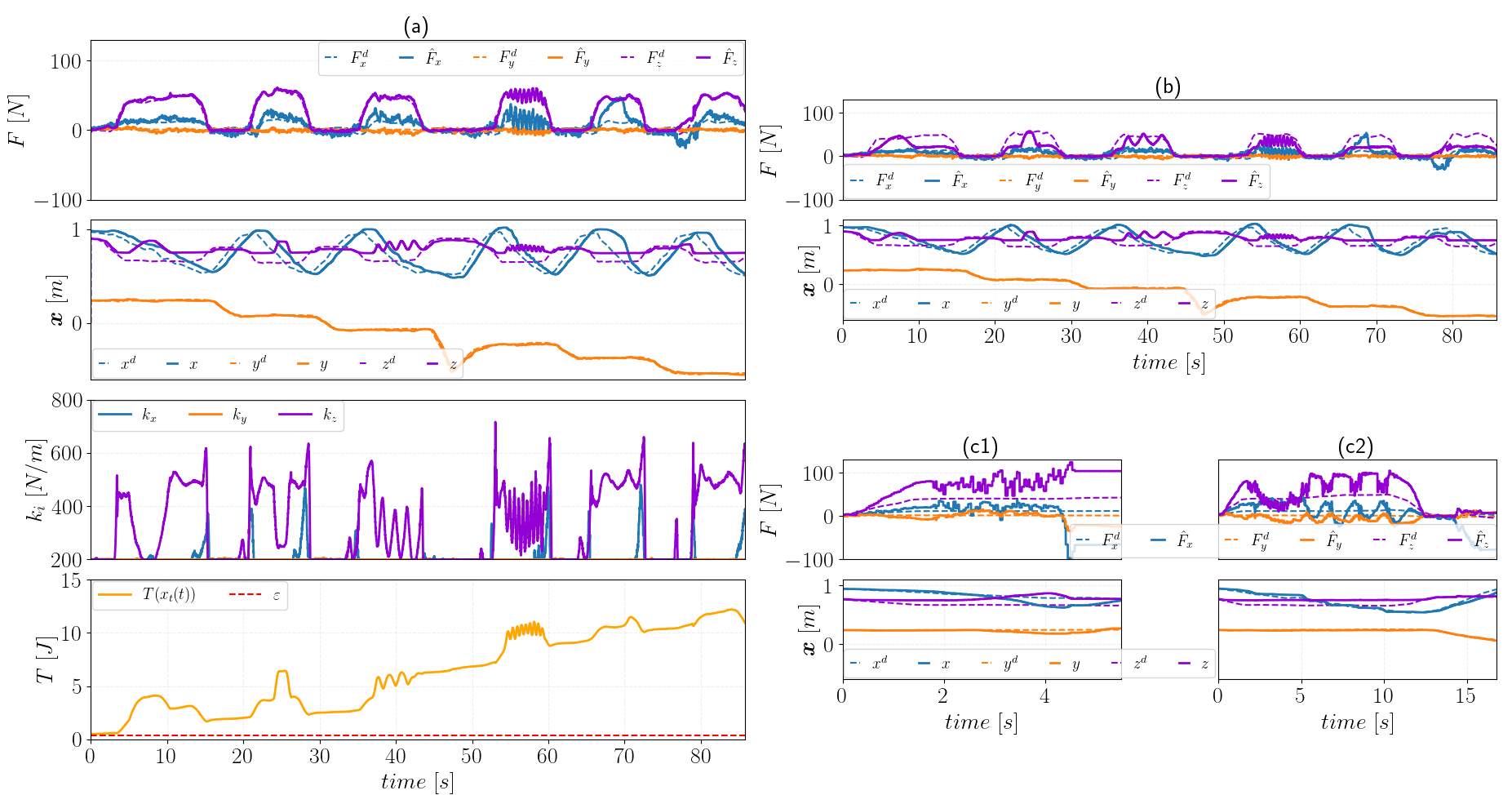}
	\caption{Results of the autonomous task repetition with external disturbances. (a). OS (b). LS (c) HS. The plotted variables are the same of Fig.~\ref{fig:repeat_experiment}. Plot (c1) shows only disturbance type 1) since the robot stopped due to violation of safety limits. Plot (c2) shows disturbances type 5) and 6), where the robot again stops due to violation of safety limits. The characterizing parameters of the first four perturbations for OS and LS respectively are (A: Amplitude; V: velocity; f: frequency): 1) A=(0.1246, 0.1230)m,  V=(0.0417,0.0464)m/s; 2) A=(0.1163, 0.1443)m, V=(0.1256,0.1163)m/s; 3)  A=(0.0758, 0.0986)m, f=(0.8645,0.6755)Hz; and 4) A=(0.0435, 0.0428)m, f=(2.0024,2.2422)Hz.}
	\label{fig:disturb_experiment}
	\vspace{-4mm}
\end{figure*}

The results of the autonomous task repetition with external disturbances for LS, HS, and OS are shown in Fig.~\ref{fig:disturb_experiment}.
In Fig.~\ref{fig:disturb_experiment}(b), the results of the autonomous task repetition with external disturbances for LS are shown. Although the robot is always compliant to environment uncertainties introduced by external disturbances, it cannot perform the task properly since insufficient force is exerted on the table. On the contrary, for HS (Fig.~\ref{fig:disturb_experiment}(c1)), as soon as the table is lifted, the interaction force exceeds the robot payload limit. The same situation occurs when a human collides along $x$ as it can be viewed in Fig.~\ref{fig:disturb_experiment}(c2) on the right. Finally, the results for OS are reported in Fig.~\ref{fig:disturb_experiment}(a). For all the types of disturbances applied along $z$, the robot can optimize the stiffness to track the desired force. The high-frequency disturbance makes the force tracking more challenging, but the performance is still acceptable. Then,  the stiffness is kept low along $x$, resulting in a compliant behavior when the human collides with the robot along that direction while interacting with the table and during free motion. In addition, the energy in the tank never falls below the threshold, and it has an overall increasing trend. At the same time, it is evident how the external disturbances affected the additional energy injected (e.g., during table lifting) and extracted (e.g., during table lowering) from the tank.

\section{DISCUSSION AND CONCLUSION}
The results described in section \ref{subsec:results_autonom_exp} show that overall OS outperforms LS and HS. Although LS keeps high compliance and robustness to external perturbations and uncertainties, it cannot achieve satisfactory cleaning performance. Instead, HS failed in performance, safety, and robustness since it renders a rigid behavior at the end-effector, resulting in high interaction forces with the environment since it rejects the external disturbances during motion. On the other hand, OS is capable of exerting the right amount of force for the cleaning task and, at the same time keeping as much compliance as possible when no interaction is required.

The framework developed in this paper allowed to easily teach a mobile manipulator to perform an interactive task. The target tasks consist of complex interaction with the environment and unconstrained motions in a larger workspace w.r.t. the reachable workspace of a fixed-base robotic arm. Moreover, the robot can robustly apply the learned interaction force while maintaining free-motion compliance. If, instead, the kinesthetic demonstrations were used to learn only desired end-effector desired position and velocity trajectories and a constant stiffness was used as it is done for HS and LS, a trade-off should have been found between ensuring compliance and high task-related performance. Even if achieving an acceptable trade-off was feasible, it would not have been possible to obtain the robustness to external perturbations that our framework accomplishes. In addition, our approach ensures the passivity of the system through an energy tank-based constraint and allows to include payload limits and stiffness bounds as constraints in the optimization problem.

Future work will include the learning procedure of all the relevant parameters of the Cartesian impedance whole-body controller, such as the motion modes and secondary tasks through an admittance-type physical interface. In this work, the task considered allowed to choose a priori some suitable controller parameters that would have ensured the successful loco-manipulation behavior of the robotic platform during the whole task execution, but in more complex environments, this is not always the case. Furthermore, the self-tuning of the weighting matrices of the QP formulation ($\boldsymbol{Q}$ and $\boldsymbol{R}$) represents another appealing future development.

%%%%%%%%%%%%%%%%%%%%%%%%%%%%%%%%%%%%%%%%%%%%%%%%%%%%%%%%%%%%%%%%%%%%%%%%%%%%%%%%

\bibliographystyle{IEEEtran}
\bibliography{biblio}

\end{document}